\documentclass[journal]{IEEEtran}

\usepackage{cite}

\ifCLASSINFOpdf
   \usepackage[pdftex]{graphicx}
\else
   \usepackage[dvips]{graphicx}
\fi

\usepackage[cmex10]{amsmath}

\DeclareMathOperator*{\argmin}{arg\,min}

\usepackage[linesnumbered, ruled, vlined]{algorithm2e}
\usepackage{algpseudocode}

\usepackage[tight,footnotesize]{subfigure}

\usepackage[font=footnotesize]{subfig}

\usepackage[table]{xcolor}

\usepackage{url}
\usepackage{diagbox}
\usepackage{array}
\usepackage{multirow}
\usepackage{booktabs}

\usepackage{color}

\begin{document}
\title{Enhance Connectivity of Promising Regions for Sampling-based Path Planning}

\author{Han Ma, Chenming Li, Jianbang Liu,
        Jiankun Wang,~\IEEEmembership{Senior Member, IEEE}
        and Max Q.-H. Meng,~\IEEEmembership{Fellow, IEEE}% <-this % stops a space
\thanks{ The work of Max Q.-H. Meng was supported in part by the National Key Research and Development Program of China under Grant 2019YFB1312400, in part by the Hong Kong Research Grants Council (RGC) General Research Fund (GRF) under Grant 14200618, and in part by the National Natural Science Foundation of China under Grant 62103181. \textit{(Corresponding authors: Jiankun Wang, Max Q.-H. Meng.)} }
\thanks{Han Ma, Chenming Li and Jianbang Liu are with the Department of Electronic Engineering, 
The Chinese University of Hong Kong, Shatin, N.T., Hong Kong SAR, China. \{hanma, licmjy, henryliu\}@link.cuhk.edu.hk.}% <-this % stops a space
\thanks{Jiankun Wang is with Shenzhen Key Laboratory of Robotics Perception and Intelligence, and the Department of Electronic and Electrical Engineering, Southern University of Science and Technology, Shenzhen 518055, China. wangjk@sustech.edu.cn.}% <-this % stops a space
\thanks{Max Q.-H. Meng is with Shenzhen Key Laboratory of Robotics Perception and Intelligence, and the Department of Electronic and Electrical Engineering, Southern University of Science and Technology, Shenzhen 518055, China, on leave from the Department of Electronic Engineering, The Chinese University of Hong Kong, Hong Kong, and also with the Shenzhen Research Institute of The Chinese University of Hong Kong, Shenzhen 518057, China. max.meng@ieee.org.}% <-this % stops a space
}

\markboth{Journal of \LaTeX\ Class Files,~Vol.~6, No.~1, January~2007}%
{Shell \MakeLowercase{\textit{et al.}}: Bare Demo of IEEEtran.cls for Journals}

\maketitle

\begin{abstract}

Sampling-based path planning algorithms usually implement uniform sampling methods to search the state space.
However, uniform sampling may lead to unnecessary exploration in many scenarios, such as the environment with a few dead ends.
Our previous work proposes to use the promising region to guide the sampling process to address the issue.
However, the predicted promising regions are often disconnected, which means they cannot connect the start and goal states, resulting in a lack of probabilistic completeness.
This work focuses on enhancing the connectivity of predicted promising regions.
Our proposed method regresses the connectivity probability of the edges in the x and y directions. 
In addition, it calculates the weight of the promising edges in loss to guide the neural network to pay more attention to the connectivity of the promising regions.
We conduct a series of simulation experiments, and the results show that the connectivity of promising regions improves significantly.
Furthermore, we analyze the effect of connectivity on sampling-based path planning algorithms and conclude that connectivity plays an essential role in maintaining algorithm performance.

\end{abstract}

\def\abstractname{Note to Practitioners}
\begin{abstract}
This work is derived from the promising region prediction for sampling-based path planning.
The sampling-based path planning methods have been widely used in robotics due to their efficiency.
To further improve the efficiency of these algorithms, sampling in the promising region predicted by a neural network is introduced into the sampling procedure.
However, the connectivity of the promising region has yet to be considered, and it will affect the performance of the algorithms in several aspects.
To demonstrate this problem, we compare the performance of the neural heuristic algorithms under different connectivity statuses in this paper. 
Furthermore, to enhance the connectivity of the predicted promising region, the novel prediction output and loss function are proposed.
The simulation results show improvements in the algorithms after utilizing our method.  
\end{abstract}


\IEEEpeerreviewmaketitle
\vspace{0.5cm}
\section{Introduction}
 
\begin{figure}[t]
  \centering
  \includegraphics[width=6.5cm]{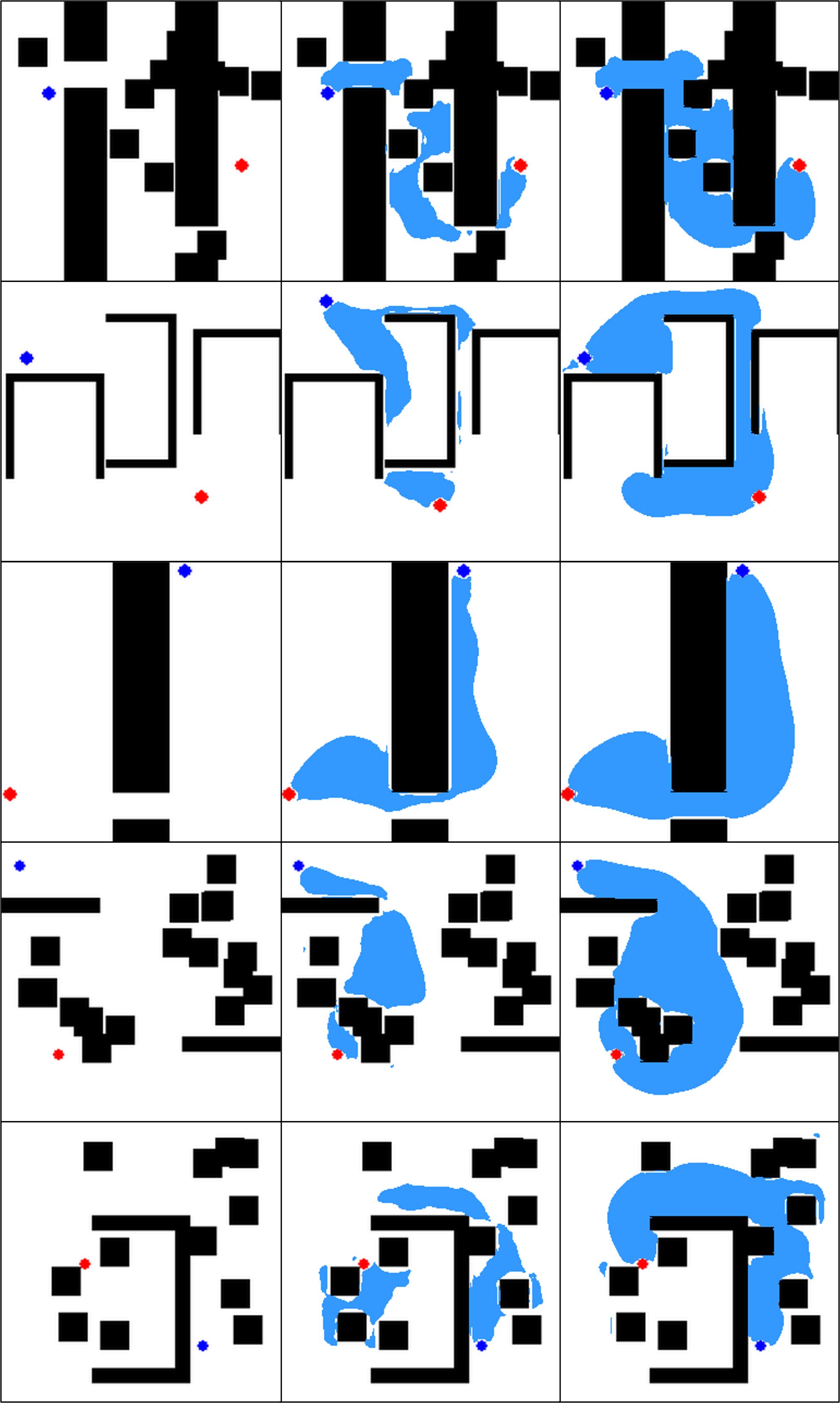}
  \caption{The blue/red dots denote the start/goal states. The sky blue regions represent the promising regions. The left column shows the original map. The middle and right columns show the disconnected and connected prediction results, respectively.}
  \label{comparison_intro}
\end{figure}

\IEEEPARstart{P}{ath} planning is one of the basic problems of robotics, and it has attracted a lot of attention in recent decades.
Path planning aims to find a feasible path that does not collide with obstacles in the environment.
Many methods have been proposed to solve the path planning problem in robotics. 
They can be roughly divided into four classes. 
The grid-based methods discretize the space into grids and then utilize the graph search algorithms like Dijkstra \cite{dijkstra1959note} and A* \cite{hart1968formal} to find the optimal path.
However, as the dimension of the search space increases, the memory and time costs increase exponentially. 
This limits the use of this method in high-dimensional search spaces.
The artificial potential field-based method \cite{khatib1986real} couples the perception feedback with the low-level control.
It constructs a force field based on the start state, the goal state, and the perception of the surrounding environment. 
The force field can drive the robot from the start to the goal without collision. 
Nevertheless, the robot may be trapped at a local minimum. 
The reward-based methods give reward to each action of the robot, such as \cite{ferguson2004focussed} \cite{bakker2005hierarchical}. 
The path planning is formulated as a Markov Decision Process (MDP), and a well-defined reward function is required to obtain a better solution.
Nevertheless, the computation load will be unacceptable when the state space becomes huge.
The sampling-based methods draw samples from the state space and construct a random graph with the samples, such as Rapidly-exploring Random Tree (RRT) \cite{lavalle2001randomized}, and Probabilistic Roadmap (PRM) \cite{kavraki1996probabilistic}. 
A feasible path is then searched on the graph.
RRT constructs a random tree incrementally from the start to the goal, which is suitable for single-query problems, while PRM constructs a random graph after the sampling process, which adapts to multi-query problems.
The sampling-based methods have been widely deployed due to their high efficiency in high-dimensional state space.
Furthermore, their variants RRT$^*$ and PRM$^*$ \cite{karaman2010incremental} are proposed for asymptotically optimal path planning.

The sampling-based methods suffer from random initial solutions and slow convergence to the optimal solution.
The reason is that they uniformly sample in the state space, which involves many redundant samples into the post-processing procedures.
This increases the burden of some already time-consuming procedures like collision checking. 
Thus, some heuristic sampling strategies are proposed to sample in the promising region where feasible or optimal solutions probably exist.
Informed RRT$^*$ \cite{gammell2014informed} is a well-known algorithm using the start, the goal, and the current solution to construct a heuristic sampling ellipsoid.
It can converge towards the optimal solution faster than RRT$^*$. 
But when the ratio of the theoretical minimum cost to the current solution cost is close to zero, informed RRT$^*$ will lose efficiency.
Deep learning is good at making predictions based on prior knowledge, which has been used in control \cite{xu2021learning} and path planning \cite{wang2020neural}.
In \cite{wang2020neural}, Neural RRT$^*$ utilizes a nonuniform sampling distribution predicted by a Convolutional Neural Network (CNN).
Generative Adversarial Network (GAN) can be also implemented to achieve nonuniform sampling \cite{ma2021conditional} \cite{zhang2021generative}.
However, they do not discuss the effect of the connectivity of the predicted promising region. 
They just use fixed heuristic sampling bias.
In this work, we focus on analyzing the connectivity of the promising region. 
Connected promising areas provide more powerful improvements for our neural heuristic algorithm, while disconnected promising areas may degrade performance.
This is explained in detail in section \ref{discussion}.

We modify the neural network's output and design connectivity loss to enhance the connectivity of promising regions and reduce the harmful effects of poor connectivity.
Specifically, our proposed method outputs the connectivity probability of the edges in the x and y directions and trains the model to pay more attention to the promising regions by computing the weight of each edge in the loss. 
In the simulation experiments, we test our method on a UNet model and compare the predicted results' connectivity rates and false-negative rates under different output formats and loss settings.
Besides, we also compare the connectivity rates and false-negative rates of the results generated by CGAN \cite{ma2021conditional}.
Fig. \ref{comparison_intro} shows the comparison between the poor connectivity and the good connectivity for the same maps. 
The middle column is selected from the predicted results with the lowest connectivity rate in section \ref{connectivity_rate_eval}. 
The right column shows the promising regions predicted by our proposed method, where the connectivity of the predicted regions is far better than the ones of the middle column.
In the right column of Fig. \ref{comparison_intro}, all the predicted results can connect the start with the goal, which will be helpful when utilizing the promising regions in sampling-based algorithms.

\subsection{Related Work}

Sampling-based path planning methods are widely deployed in various robot systems due to their efficiency.
RRT \cite{lavalle2001randomized}, and PRM \cite{kavraki1996probabilistic} both are fundamental and well-known sampling-based methods, and both of them guarantee probabilistic completeness.
However, no asymptotic optimality is provided by RRT or PRM. 
RRT$^*$ and PRM$^*$ are the variants of RRT and PRM, which provide asymptotic optimality as the number of samples goes infinite.
Nevertheless, the converge rates of RRT$^*$ and PRM$^*$ are expected to be improved.
Fast Marching Tree (FMT$^*$) \cite{janson2015fast} combines the features of RRT and PRM and performs a 'lazy' dynamic programming recursion.
The numerical experiments demonstrate that FMT$^*$ converges faster than RRT$^*$ and PRM$^*$.
Informed RRT$^*$ is another try to improve the converge rate of RRT$^*$ by sampling in a designed hyper ellipsoid. 
Furthermore, based on informed RRT$^*$, Batch Informed Trees (BIT$^*$) \cite{gammell2015batch} is proposed.
BIT$^*$ finds better solutions than RRT$^*$, informed RRT$^*$ and FMT$^*$ with a faster converge rate. 
Its improvements benefit from processing samples in batches and sampling in a heuristic hyper ellipsoid.
Adaptively Informed Trees (AIT$^*$) \cite{strub2020adaptively} deriving from BIT$^*$ uses a reverse tree to estimate the cost-to-go. 
The reverse tree performs a lazy search with Lifelong Planning A$^*$ (LPA$^*$) \cite{koenig2004lifelong}.
It finds the solution as fast as RRT-Connect \cite{kuffner2000rrt} while keeping the asymptotic optimality.
Mandalika et al. \cite{mandalika2021guided} design a local densification method to make the informed set more efficient. 
In their method, local subsets defined by beacon nodes are leveraged to guide sampling, which increases the utilization of the informed set.

The above methods counting on manually designed heuristic lose efficacy in some extreme environments, and besides, the sampling complexity is desired to be reduced.
Deep learning methods' flourishing motivates researchers to utilize them to solve path planning problems.
\cite{zhang2018learning} presents a policy-based search method to learn the implicit sampling distribution that is implemented in rejecting sampling manner.
\cite{ichter2020learned} proposes critical PRM, which leverages neural networks to identify the critical points in the maps for path planning. 
Ichter et al. \cite{ichter2018learning} learn a nonuniform sampler with a Conditional Variational Autoencoder (CVAE) \cite{doersch2016tutorial}. 
The data are collected from previous successful motion plans.
Kumar et al. \cite{kumar2019lego} propose a promotion of \cite{ichter2018learning}, called Leveraging Experience with Graph Oracles (LEGO).
LEGO trains a CVAE model on a diverse shortest paths set, and then, the trained model predicts the bottleneck nodes.
Jenamani et al. \cite{jenamani2020robotic} further improve LEGO.
\cite{jenamani2020robotic} presents a method to locate the bottleneck regions and proposes two methods using local sampling to utilize the bottleneck locations. 
Qureshi et al. \cite{qureshi2018deeply} put forward a neural network-based adaptive sampler for sampling-based motion planning, named DeepSMP.
DeepSMP consists of an environment encoder encoding raw point cloud data and a random sampler based on dropout layers. 
The training data are generated by RRT$^*$.
In \cite{qureshi2020motion}, Qureshi et al. present Motion Planning Networks (MPNet), which contains an encoder network and a planning network.
Like DeepSMP, the encoder network embeds the environment information (e.g., point cloud from depth camera or LIDAR) to a latent space.
Then, the planning network takes the current state, goal state, and latent variable and outputs a new state.
In such a way, MPNet can solve a planning problem with near-optimal heuristics.
In \cite{khan2020graph}, Khan et al. use Graph Neural Networks (GNNs) to encode the topology of the state space,
and two different methods are proposed, one for constant graph and others for the incremental graph.
\cite{wang2020neural} and \cite{ma2021conditional} predict promising region to guide the sampling.
However, the effects of the connectivity quality of the promising region are not stated, and the connectivity of the promising region is not enhanced.

Some work tries to improve the connectivity of the prediction results in the computer vision field where the same kind of pixels is considered to be connected.   
In \cite{kampffmeyer2018connnet}, to segment an image more accurately, Kampffmeyer et al. insert a non-local block into the encoder and take eight directions' connectivity of the pixels as labels.
\cite{turaga2009maximin} presents a loss minimizing the topological error on the predicted connectivity, and the method is improved in \cite{funke2018large}.
These methods regard the image as a graph in the loss computation, which works well in their tasks. 
In path planning problems, the topological features of the map data are much more important than the texture features. 
Thus, we propose a new output format for the model and design a new loss function to achieve better connectivity of the predicted promising region.

\subsection{Original Contributions}
The contributions of this paper include:
\begin{itemize}
  \item we propose a UNet based model that outputs the connectivity probability of the edges in the x and y directions and design the connectivity loss to guide the learning of the model, 
  which enhances the connectivity of the predicted promising regions significantly;
  \item we evaluate the effects of different connectivity statuses on the sampling-based planning algorithms;
  \item besides, we discuss the performance of the sampling-based methods under different sampling biases in the promising regions and demonstrate the practicability of our neural heuristic methods with simulations in the Robot Operating System (ROS).
\end{itemize} 

The rest of this paper is organized as follows. 
We formulate the problem and introduce the background methods briefly in section \ref{preliminiaries}.
Section \ref{methodology} presents the structure of the neural network model and explains the details of the designed loss functions.
In section \ref{simulation_results}, we present the simulation results of our method. 
Section \ref{discussion} discusses the effects of different connectivity statuses and heuristic sampling biases on the performance of the sampling-based path planning algorithms, and section \ref{discussion} also discusses the performance of neural heuristic methods in robot navigation tasks. 
In the end, we draw a conclusion and discuss the future work in section \ref{conclusions}.

\section{Preliminaries}
\label{preliminiaries}
Firstly, we formulate the path planning problem in section \ref{path_planning_problem}. Secondly, section \ref{nh_rrts} overviews neural heuristic RRT and RRT$^*$ algorithm.
Thirdly, we describe the map representations in section \ref{map_repr}. 
At last, section \ref{cbpt_intro} gives a brief introduction of the Canonical Binary Partition Tree (CBPT) \cite{cousty2018hierarchical} used in our loss computation.

\subsection{Path Planning Problem}
\label{path_planning_problem}

Let $\mathcal{X} \subset \mathcal{R}^{n}$ be the state space of the planning problem. 
$\mathcal{X}_{free} \subset \mathcal{X}$ denotes the collision-free subspace of the state space.
$\mathcal{X}_{obs} = \mathcal{X} \setminus \mathcal{X}_{free}$ denotes the obstacles in the state space.
$x_1, x_2 \in \mathcal{X}$ are any two different states in $\mathcal{X}$. 
The Euclidean distance between $x_1$ and $x_2$ can be computed by $L_2$ norm, $||x_1 - x_2||_2$.
The $r$-radius ball centers at $x$ is denoted as $\mathcal{B}(x, r)$.
Let $x_s \in \mathcal{X}_{free}$ and $x_g \in \mathcal{X}_{free}$ be the start state and goal state, respectively.
Let $\sigma : [0, T] \mapsto \mathcal{X}_{free}$ be a feasible solution to the planning problem. 
Then, $\sigma$ is a sequence of states, and we have $\sigma(0) = x_s$, $\sigma(T) \in \mathcal{B}(x_g, r_g)$ and $\sigma(t) \in \mathcal{X}_{free}, t \in [0, T]$, where $r_g$ is the predefined threshold. 
Let $\Sigma$ be the set of feasible solutions and $\sigma \in \Sigma$.
The cost of path $\sigma$ is defined as $c(\sigma)= \Sigma^{T}_{t=1} ||\sigma(t) - \sigma(t-1)||_2$.
The optimal path planning algorithm can be formulated as below:
\begin{equation}
  \begin{aligned}
    \sigma^* = \argmin_{\sigma} \quad & c(\sigma)\\
    \textrm{s.t.} \quad & \sigma \in \Sigma\\
  \end{aligned}
\end{equation}
where $\sigma^{*}$ is the optimal solution.

\subsection{Neural Heuristic RRT and RRT$^*$}
\label{nh_rrts}

\begin{algorithm}[t]
  \caption{NH-RRT/RRT*}
  \label{nh_rrt}
  $V \gets \emptyset$, $E \gets \emptyset$\;
  \For{$i=1$ \KwTo $n$}{
    \Repeat{$FreeState(x_{new})$}{
        \If {$Random(0, 1) > h_{b}$}{
          $x_{rand} \gets Sample_u()$\;
        }
        \Else{
          $x_{rand} \gets Sample_h()$\;
        }
        $x_{nearest} \gets Nearest(x_{rand})$\;
        $x_{new} \gets Steer(x_{nearest}, x_{rand})$\;
    }
  
    \If {$RRT$}{
      \If {$FreeEdge(x_{nearest}, x_{new})$}{
        $V \gets V \cup x_{new}$\;
        $E \gets E \cup \{x_{nearest}, x_{new}\}$\;
      }
    }
    \ElseIf{$RRT^*$}{
      $x_{parent} \gets ChooseParent(x_{new})$\;
      $V \gets V \cup x_{new}$\;
      $E \gets E \cup \{x_{parent}, x_{new}\}$\;
      $Rewire()$\;
    }
    \If{$Terminate()$}{
      $Break$\;
    }
  }
\end{algorithm}

The evaluation and demonstration of our method are based on neural heuristic RRT and RRT$^*$, abbreviated to NH-RRT and NH-RRT$^*$.
This section gives a brief introduction to NH-RRT and NH-RRT$^*$. 
RRT consists of several subroutines, including random sampling, finding the nearest state, collision checking of states and edges, and steering. 
RRT$^*$ is a modification of RRT, and it adds two subroutines to obtain asymptotic optimality.
One is choosing a parent who finds a parent state for the new state in its near neighbor, and the other is rewiring which rewires the neighbor edges of the new state to get smaller cost-to-comes.
Both RRT and RRT$^*$ grow a random search tree $\mathcal{T}$ incrementally as the sampling process goes on.
In NH-RRT and NH-RRT$^*$, the neural heuristic is introduced into the random sampling procedure to sample in the promising region.
Let $V$ be the set of the vertices in $\mathcal{T}$ and $E$ be the set of edges between the vertices, i.e., $\mathcal{T} = \{V, E\}$.
The uniform random sampling and heuristic random sampling is denoted as $Sample_{u}()$ and $Sample_{h}()$, respectively.
$FreeState(x)$ returns the collision status of $x$, and $FreeEdge(x_1, x_2)$ returns the collision status of edge $\overline{x_1 x_2}$.
$ChooseParent(x)$ returns the best parent state $x_{parent}$ of $x$, and $\{x_{parent}, x\}$ is collision free. 
$Rewire()$ represents the rewiring subroutine. 
$Nearest(x)$ returns the nearest neighbor of $x$ in $\mathcal{T}$.
The whole procedure of NH-RRT/RRT$^*$ is summarized in \textbf{Algorithm} \ref{nh_rrt} where $x_{rand}$, $x_{new}$, $x_{nearest}$ and $x_{parent}$ represent random state, new state, the nearest state and the parent state, respectively.
$Random(0, 1)$ generates a random number distributing uniformly in $[0, 1]$. 
The heuristic sampling bias is denoted as $h_b$. 
In section \ref{discussion}, to demonstrate the effects of $h_b$ on the algorithms, we change the value of $h_b$ under different connectivity statuses.

\subsection{Map Representations}
\label{map_repr}

\begin{figure}[!t]
    \centering
    \includegraphics[width=8.0cm]{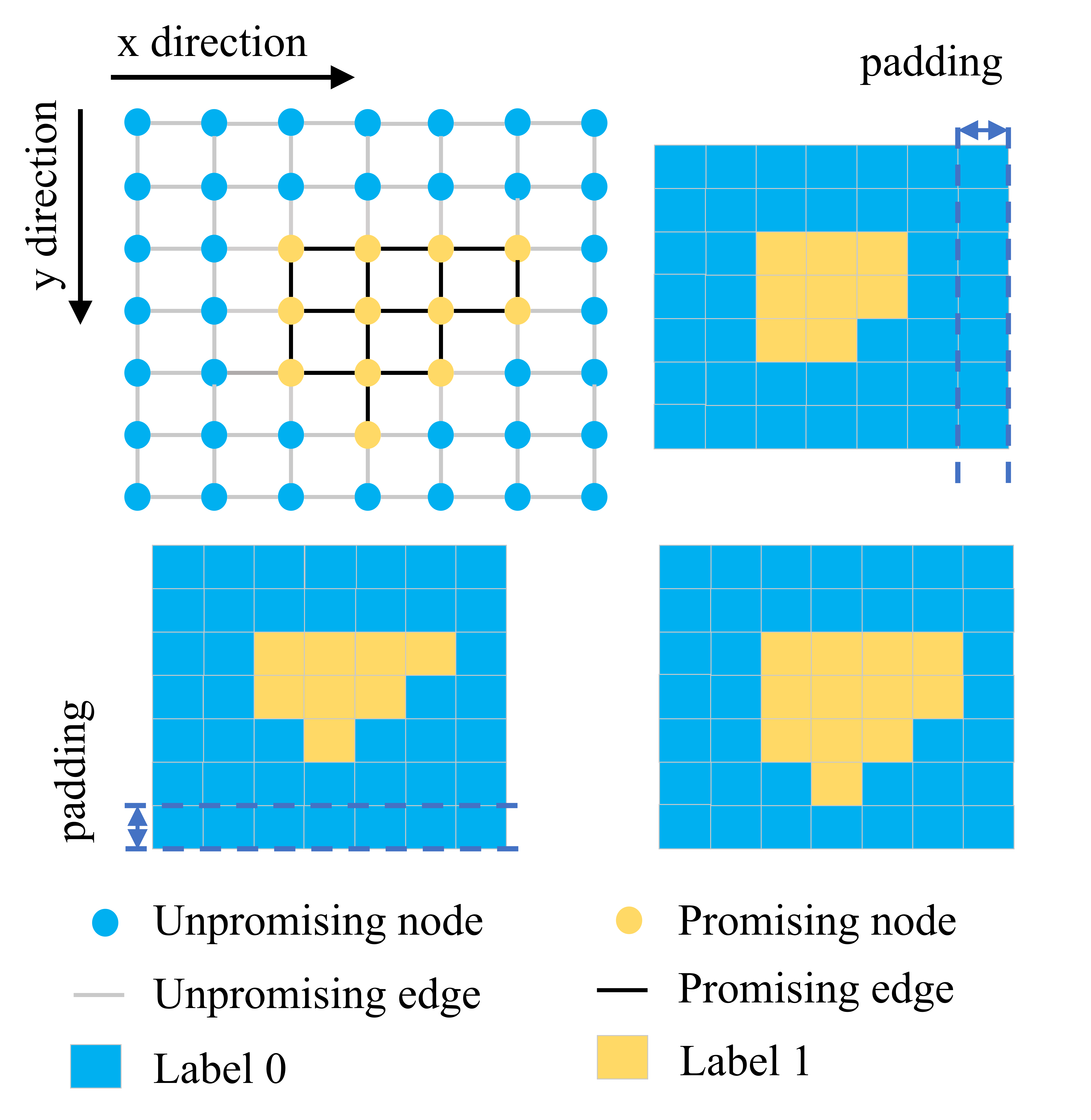}
    \caption{Output representation and the corresponding labels.}
    \label{connectivity}
\end{figure}
The map in robot path planning contains all the information of free space and obstacle space.
It can be represented by a lattice, and in the implementation, we store it as the image format shown in Fig. \ref{comparison_intro}.
The white region denotes free space, while the black region denotes obstacle space.
In the promising region prediction task, each pixel in the map data represents a candidate node.
The objective is to select all the promising nodes that near-optimal solutions are more likely to go through.
In the output of the neural network model, to place extra emphasis on the connectivity, instead of regarding the map as an image, we represent the map with a graph.
As is shown in Fig. \ref{connectivity}, the graph consists of the candidate nodes and the edges between them. 
The edges between the promising nodes are promising, while other edges are unpromising.
In the left top of Fig. \ref{connectivity}, the blue and yellow nodes in the lattice represent unpromising and promising nodes, respectively.
The horizontal direction is defined as the x-direction, and the vertical direction is defined as the y-direction.
We label the edges in two directions instead of labeling the nodes. 
The promising edges are labeled ones, while the unpromising ones are labeled zeroes.
To keep the size consistent with the original map, we add a column padding and a row padding to the labels in x and y directions, and the paddings are labeled zeroes.
The left bottom and right top of Fig. \ref{connectivity} show the corresponding labels of each direction, as opposed to the node label at the right bottom of Fig. \ref{connectivity}.
We find that with this kind of label and the proposed connectivity loss, compared with labeling the nodes, the connectivity of the prediction results is significantly improved.

\subsection{Canonical Binary Partition Tree}
\label{cbpt_intro}

\begin{figure}[!t]
  \centering
  \includegraphics[width=6.5cm]{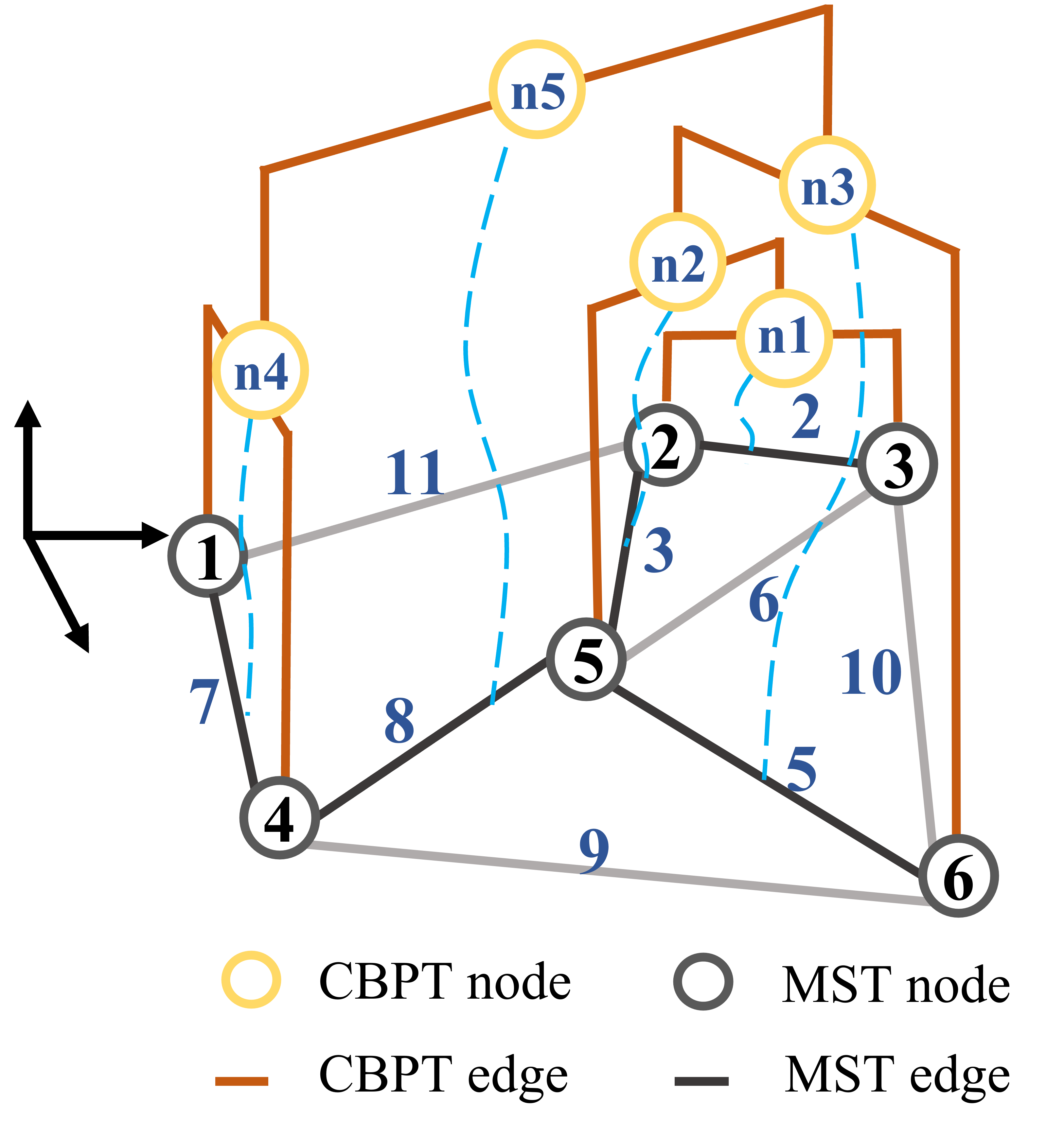}
  \caption{This figure shows that a CBPT is constructed upon a MST.
  The dotted blue curves denote the associations between nodes of the CBPT and the MST.}
  \label{cbpt_fig}
\end{figure}

\begin{figure*}[!ht]
  \centering
  \includegraphics[width=16.5cm]{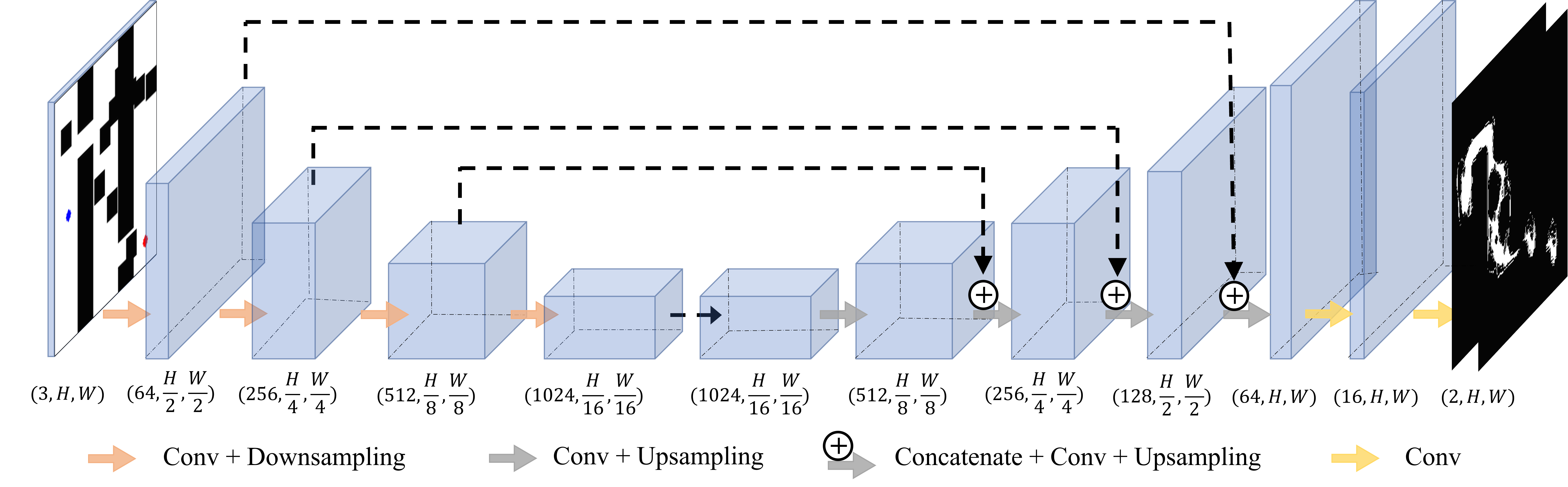}
  \caption{Illustration of the structure of the neural network model.}
  \label{nn_model}
\end{figure*}

Let $V = \{v_0, v_1, ..., v_n\}$ be a node set and $E = \{e_0, e_1, ..., e_k\}$ be an ordered edge set, where $e_i \prec e_{i+1}$ means $e_i$ has smaller weight than $e_{i+1}$ and $i \in \{0, 1, ..., k-1\}$.
Given a weighted graph $G= \{V, E\}$, a CBPT is constructed according to the weights of the edges $e_{i} \in E$.
Then, a Minimum Spanning Tree (MST) can be established based on the order of $E$. Establishing an MST is a merging process of the nodes on $G$.
Let $B$ be the edge set of the MST. According to the order of $E$, the MST algorithm tries to add an edge $e_i$ to $B$ in turn.
If at least one of the extremities of $e_i$ is not included in any of the edges in $B$, $e_i$ is added to $B$. Otherwise, $e_i$ is abandoned. 
This process is repeated until $|B| = n - 1$, which means the MST corresponding to $G$ is established successfully, and $|\cdot |$ denotes the cardinality of a set.
A CBPT records the merging steps of an MST, and each node of CBPT is related to an edge of MST. 
Fig. \ref{cbpt_fig} presents an example of the relationship between CBPT and MST. 
Intuitively, the node related to an edge with a smaller weight is closer to the leaf node of CBPT, and otherwise, it is closer to the root node of CBPT.
The root node of CBPT is related to the edge with the most significant weight in MST.

\section{Methodology}
\label{methodology}

In this section, we first introduce the structure of the neural model in section \ref{neural_model}. 
Then, we describe the proposed loss function in detail in section \ref{loss_function}.
At last, section \ref{promising_region_computation} formulates the promising region computation.

\subsection{Neural Network Model}
\label{neural_model}

The input data is a 3-channel 2D map image with the start and goal states. 
As illustrated in Fig. \ref{comparison_intro}, black color represents the obstacle space, and white color represents the free space.
The red and blue dots represent the start and the goal, respectively.
For the sake of concise, we denote the input map with size $(3, H, W)$ as $\mathcal{M}$.
Before being fed to the neural model, $\mathcal{M}$ is normalized. 
The output of the neural model is a 2-channel tensor $\mathcal{P}$ with size $(2, H, W)$.
Each channel of $\mathcal{P}$ is the probabilistic distribution indicating the connectivity probability of the edges in a specific direction.
Let $\mathcal{P}^x$ and $\mathcal{P}^y$ denote the two channels of $\mathcal{P}$. 
Then, $\mathcal{P}^x$ is the edge connectivity probabilistic distribution in the x-direction, and $\mathcal{P}^y$ is the edge connectivity probabilistic distribution in the y-direction. 
Each element of $\mathcal{P}$, i.e., $\mathcal{P}_{i, j} \in [0, 1]$ indicates the probability that the corresponding edge connects two promising nodes.
The higher value of $\mathcal{P}_{i, j}$ means the higher probability the corresponding edge lies in the promising region.

Our promising region prediction task is similar to semantic segmentation. 
The objective of both tasks is to select areas worthy of attention. 
Thus, inspired by the popular methods for semantic segmentation, we use a CNN-based model to obtain the features of the map. 
Multiscale features should be considered to predict a more accurate promising region.
We utilize the UNet structure to satisfy this requirement that can fuse multiscale features through skip connections.
Fig. \ref{nn_model} shows the structure of the UNet model, and three skip connections are utilized to fuse the features in the encoder with those in the decoder.
There are four layers in total in the encoder, and each layer consists of a residual block, including a downsampling process with a factor of 2.
Symmetrical to the encoder, the decoder comprises four upsampling layers with a factor of 2.
Downsampling is achieved by setting the convolution stride to 2, and upsampling is achieved through bilinear interpolation.
If we denote a feature map with $k$ channels as $F_k$, the encoder used can be described as: $F_{3} - F_{64} - F_{256} - F_{512} - F_{1024}$, and the decoder used can be described as: $F_{1024} - F_{512} - F_{256} - F_{128} - F_{64} - F_{16} - F_{2}$.
Assuming that the size of the input is $(H, W)$, the sizes of the features maps in the encoder are $(\frac{H}{2}, \frac{W}{2})$, $(\frac{H}{4}, \frac{W}{4})$, $(\frac{H}{8}, \frac{W}{8})$, $(\frac{H}{16}, \frac{W}{16})$.
The sizes of the features maps in the decoder are $(\frac{H}{16}, \frac{W}{16})$, $(\frac{H}{8}, \frac{W}{8})$, $(\frac{H}{4}, \frac{W}{4})$, $(\frac{H}{2}, \frac{W}{2})$, $(H, W)$, $(H, W)$.

\subsection{Loss Functions}
\label{loss_function}
% BCE loss
The objective of the promising region prediction task is to distinguish the promising nodes from the unpromising ones in the input map $\mathcal{M}$.
Binary cross entropy is widely used to evaluate the difference between two distributions for a given random variable.
We expect our model to output the same distribution as the target distribution.
Henceforth, Binary Cross Entropy (BCE) loss is performed in the loss function of our task. 
As is specified in section \ref{map_repr}, we label the edges in x and y directions rather than label the nodes.
The BCE loss can be divided into two branches, $L_{bce}^x$ and $L_{bce}^y$. 
(\ref{l_bce_x}) and (\ref{l_bce_y}) specify the computation of $L_{bce}^x$ and $L_{bce}^y$, where $\mathcal{P}$ is the target distribution and $\hat{\mathcal{P}}$ is the predicted distribution.
\begin{equation}
  \label{l_bce_x}
  L_{bce}^x (\mathcal{P}^x, \hat{\mathcal{P}}^x)=  \sum_{i=1}^{H}\sum_{j=1}^{W} \mathcal{P}^x_{i,j} log(\hat{\mathcal{P}}^x_{i,j}) \\ + (1 - \mathcal{P}^x_{i,j}) log(1 - \hat{\mathcal{P}}^x_{i,j})
\end{equation}
\begin{equation}
  \label{l_bce_y}
  L_{bce}^y (\mathcal{P}^y, \hat{\mathcal{P}}^y)=  \sum_{i=1}^{H}\sum_{j=1}^{W} \mathcal{P}^y_{i,j} log(\hat{\mathcal{P}}^y_{i,j}) \\ + (1 - \mathcal{P}^y_{i,j}) log(1 - \hat{\mathcal{P}}^y_{i,j})
\end{equation}

The overall BCE loss $L_{bce}^{xy}$ consists of $L_{bce}^x$ and $L_{bce}^y$.
$L_{bce}^{xy}$ penalizes wrong classification results, the greater the difference between $\mathcal{P}$ and $\hat{\mathcal{P}}$, the larger the value of $L_{bce}^{xy}$. 
\begin{equation}
  L_{bce}^{xy} (\mathcal{P}, \hat{\mathcal{P}}) = L_{bce}^x (\mathcal{P}^x, \hat{\mathcal{P}}^x) + L_{bce}^y (\mathcal{P}^y, \hat{\mathcal{P}}^y)
\end{equation}

% DICE loss
In our task, the promising region may only occupy a small part of the whole map.
This may result in the neural network model getting trapped in a local minimum.
The model's output will be heavily biased towards the unpromising area, which means the predicted promising region will be inaccurate, and its connectivity will not be guaranteed. 
Dice loss ($L_{dice}$) is an objective function based on the dice coefficient. 
Intuitively, the larger the dice coefficient, the more accurate the prediction result is relative to the ground truth.
To derive $L_{dice}^{xy}$, we first need to compute the dice coefficient of the two channels.
The intersection of each channel is defined in (\ref{intersection_x}) and (\ref{intersection_y}), respectively.
The cardinality of each channel is defined in (\ref{cardinality_x}) and (\ref{cardinality_y}), respectively.
$L_{dice}^{xy}$ is one minus the dice coefficient as defined in (\ref{l_dice}).
\begin{equation}
  \label{intersection_x}
  |\mathcal{P}^x \cap  \hat{\mathcal{P}}^x| = \sum_{i=1}^{H}\sum_{j=1}^{W} \mathcal{P}^x_{i,j} * \hat{\mathcal{P}}^x_{i,j}
\end{equation}
\begin{equation}
  \label{cardinality_x}
  |\mathcal{P}^{x2}| + |\hat{\mathcal{P}}^{x2}| = \sum_{i=1}^{H}\sum_{j=1}^{W} \hat{\mathcal{P}}^x_{i,j} * \hat{\mathcal{P}}^x_{i,j} 
  + \sum_{i=1}^{H}\sum_{j=1}^{W} \mathcal{P}^x_{i,j} * \mathcal{P}^x_{i,j}
\end{equation}
\begin{equation}
  \label{intersection_y}
  |\mathcal{P}^y \cap  \hat{\mathcal{P}}^y| = \sum_{i=1}^{H}\sum_{j=1}^{W} \mathcal{P}^y_{i,j} * \hat{\mathcal{P}}^y_{i,j}
\end{equation}
\begin{equation}
  \label{cardinality_y}
  |\mathcal{P}^{y2}| + |\hat{\mathcal{P}}^{y2}| = \sum_{i=1}^{H}\sum_{j=1}^{W} \hat{\mathcal{P}}^y_{i,j} * \hat{\mathcal{P}}^y_{i,j} 
  + \sum_{i=1}^{H}\sum_{j=1}^{W} \mathcal{P}^y_{i,j} * \mathcal{P}^y_{i,j}
\end{equation}
\begin{equation}
  \label{l_dice}
  L_{dice}^{xy}(\mathcal{P}, \hat{\mathcal{P}}) = 1 - \frac{2 |\mathcal{P}^x \cap  \hat{\mathcal{P}}^x| + 2 |\mathcal{P}^y \cap  \hat{\mathcal{P}}^y| }{|\mathcal{P}^{x2}| + |\hat{\mathcal{P}}^{x2}| + |\mathcal{P}^{y2}| + |\hat{\mathcal{P}}^{y2}|}
\end{equation}

% connectivity loss
To enhance the connectivity of the promising region prediction, we design the connectivity loss denoted as $L_c$. 
$L_c$ is only performed on the promising region indicated by the ground truth. 
Let the graph corresponding to $\mathcal{M}$ be $\mathcal{G}$, and $\mathcal{G} = \{ \mathcal{E}, \mathcal{V}, f_p \}$, where $\mathcal{E}$ and $\mathcal{V}$ are the edge set and the node set, respectively, and $f_p$ judges whether a node is promising or not, i.e., $f_p: v \to \{0, 1 \}, v \in \mathcal{V}$.
Let $e_k$ be an edge in $\mathcal{E}$, where $k \in \{1, 2, 3, ...., 2HW-H-W \}$ is the index of the edge.
As defined in section \ref{neural_model}, every edge in $\mathcal{E}$ is corresponding to an element of $\mathcal{P}$.
Here, we define a mapping $p: \mathcal{E} \to \mathcal{P}$ to map $\mathcal{E}$ to $\mathcal{P}$.
Given an arbitrary edge $e \in \mathcal{E}$, $\delta: \mathcal{E} \to \{0, 1\}$ indicating whether $e$ lies in the promising region or not, which is defined below:
\begin{equation}
	\delta(e) = \begin{cases}
	1, & if \ f_p(e(0))=1 \ and \ f_p(e(1))=1,\\
	0, & otherwise,	
	\end{cases}
\end{equation}
where $e(0)$ and $e(1)$ are the two nodes connected by $e$.
To guide the neural network model to pay more attention to the promising region and make the back propagation more effective, the weight of each edge in $\mathcal{E}$ should be computed. 
The weight of $e \in \mathcal{E}$ is defined in (\ref{edge_weight}), where $P(u, v, MST(\hat{\mathcal{P}}))$ means the path connecting $u$ with $v$ in the $MST$ of $\mathcal{G}$,
and the edge weights of $\mathcal{G}$ is given by the predicted connectivity probability $\hat{\mathcal{P}}$.
\begin{equation}
  \begin{aligned}
  \label{edge_weight}
	w(e)= & |\{ (u, v) \ | f_p(u) =1 \ and \ f_p(v) = 1, \\
        & \delta(e)=1, \ e \in P(u, v, MST(\hat{\mathcal{P}})) \}|
  \end{aligned}
\end{equation}
According to (\ref{edge_weight}), the weight can be expressed as (\ref{edge_weight_int}) more concisely.
\begin{equation}
  \label{edge_weight_int}
	w(e)=|V_{T_0}||V_{T_1}|
\end{equation}
where $T_0$ and $T_1$ are two subtrees of $MST(\hat{\mathcal{P}})$ merged by $e$, i.e., $T_0 \subset MST(\hat{\mathcal{P}})$, $T_1 \subset MST(\hat{\mathcal{P}})$, and $e$ lies in the promising region, i.e., $\delta(e)=1$.
$|\cdot|$ in (\ref{edge_weight}) and (\ref{edge_weight_int}) denotes the cardinality of a set.
$V_{T_0}$ is the set of promising nodes in $T_0$, and $V_{T_1}$ is the set of promising nodes in $T_1$. 
$|V_{T_0}|$ and $|V_{T_1}|$ can be computed by CBPT algorithm mentioned in section \ref{cbpt_intro}.
Intuitively, $w(e)$ is determined by the promising nodes pairs that $e$ connects. 
If $e$ connects more pairs of the promising nodes, $e$ is more critical in the promising region, and thus, $w(e)$ has a higher value. 
As $L_c$ only considers the promising region in $\mathcal{M}$, $L_c$ can be computed through the error between the predicted connectivity probability and the target connectivity probability (i.e., $\delta(e_k)=1$).
The computation of $L_c$ is summarized as below:
\begin{equation}
	L_{c}(\mathcal{P}, \hat{\mathcal{P}})= \sum_{e_k \in MST(\hat{\mathcal{P}}), \delta(e_k)=1} w(e_k)||1 - p(e_k) ||^2.
\end{equation}
As the value of $w(e)$ may be very large, and we want to combine the efficacy of the aforementioned two losses,
we normalize $L_{c}$ by dividing the sum of all edge weights.
\begin{equation}
	L_{c}(\mathcal{P}, \hat{\mathcal{P}})= \frac{\sum_{e_k \in MST(\hat{\mathcal{P}}), \delta(e_k)=1} w(e_k)||1 - p(e_k) ||^2}{\sum_{e_k \in MST(\hat{\mathcal{P}}), \delta(e_k)=1} w(e_k)}
\end{equation}

The overall loss defined in (\ref{overall_loss}) comprises BCE loss, Dice loss, and connectivity loss. 
$L_{bce}^{xy}$ penalizes inaccurate predictions in both promising and unpromising areas. 
$L_{dice}^{xy}$ guides the network to avoid the local optimum in order to predict a more accurate promising region.
Furthermore, $L_c$ focuses on predicting the promising region, allowing the network to pay attention to the area that causes disconnection.
Combining the capabilities of the three losses enables the trained network to ensure the connectivity of the network prediction results in long distances and narrow passages.
\begin{equation}
  \label{overall_loss}
	L(\mathcal{P}, \hat{\mathcal{P}}) = L_{bce}^{xy}(\mathcal{P}, \hat{\mathcal{P}}) + L_{dice}^{xy}(\mathcal{P}, \hat{\mathcal{P}}) + L_{c}(\mathcal{P}, \hat{\mathcal{P}})
\end{equation}

\subsection{Promising Region Computation}
\label{promising_region_computation}
As our model outputs the connectivity probability of the edges, we need to convert it into a promising region prediction map.
The key to the conversion problem is to judge whether nodes on the map are promising by the connectivity status of the edges.
As is shown in Fig. \ref{connectivity}, one promising node at least has one promising edge. 
In the neural network's output $\hat{\mathcal{P}}$, an edge is promising means that the value of the corresponding element of $\hat{\mathcal{P}}$ is larger than the unpromising ones. 
Here, we define a threshold $t$ to evaluate the connectivity of an edge. If $\hat{\mathcal{P}}_{i,j} > t$, the corresponding edge of $\hat{\mathcal{P}}_{i,j}$ is promising.
In the simulation experiments, we set $t$ to $0.09$ empirically.
Let $\mathcal{P}^f$ denote the final promising region prediction result.
Then, the computation of $\mathcal{P}^f$ can be formulated as (\ref{final_pred_result}).
\begin{equation}
  \label{final_pred_result}
  \mathcal{P}^f_{i,j} = \begin{cases}
    1, & if \ \frac{\hat{\mathcal{P}}^x_{i,j} + \hat{\mathcal{P}}^y_{i,j}}{2} > t,\\
    0, & otherwise.	
  \end{cases}
\end{equation}

\section{Simulation Results}
\label{simulation_results}

In this section, we demonstrate the performance of the proposed method by comparing it with different output formats, and loss function designs and also our previous method \cite{ma2021conditional}, on the connectivity rates of the test set. 
The organization of this section is as follows.
In section \ref{network_training_details}, we describe the hyperparameter setup and training details of the neural network, and we also introduce the generation of the dataset used in the training.
In section \ref{connectivity_rate_eval}, we first describe the four methods that participate in the comparison and then evaluate the connectivity rates of these methods on the test set.

\subsection{Network Training Details}
\label{network_training_details}

The neural network model is trained on NVIDIA GeForce RTX3080 with Pytorch. 
Mini-batch Stochastic Gradient Descent (SGD) optimizer is used in the training.
The momentum and weight decay of the optimizer are set to $0.9$ and $1.0 \times 10^{-4}$, respectively.
$lr$ is updated according to the rule below: 
\begin{equation}
lr = lr \times (1.0-\frac{I_{num}}{I_{max\_iter}})^{0.9},
\end{equation}
where $I_{num}$ is current iteration steps, and $I_{max\_iter}$ is the total iteration steps.
We train the model for 30 epochs with batch size 30 with all methods involved in the comparison.

\begin{figure}[!t]
  \centering
  \includegraphics[width=8.5cm]{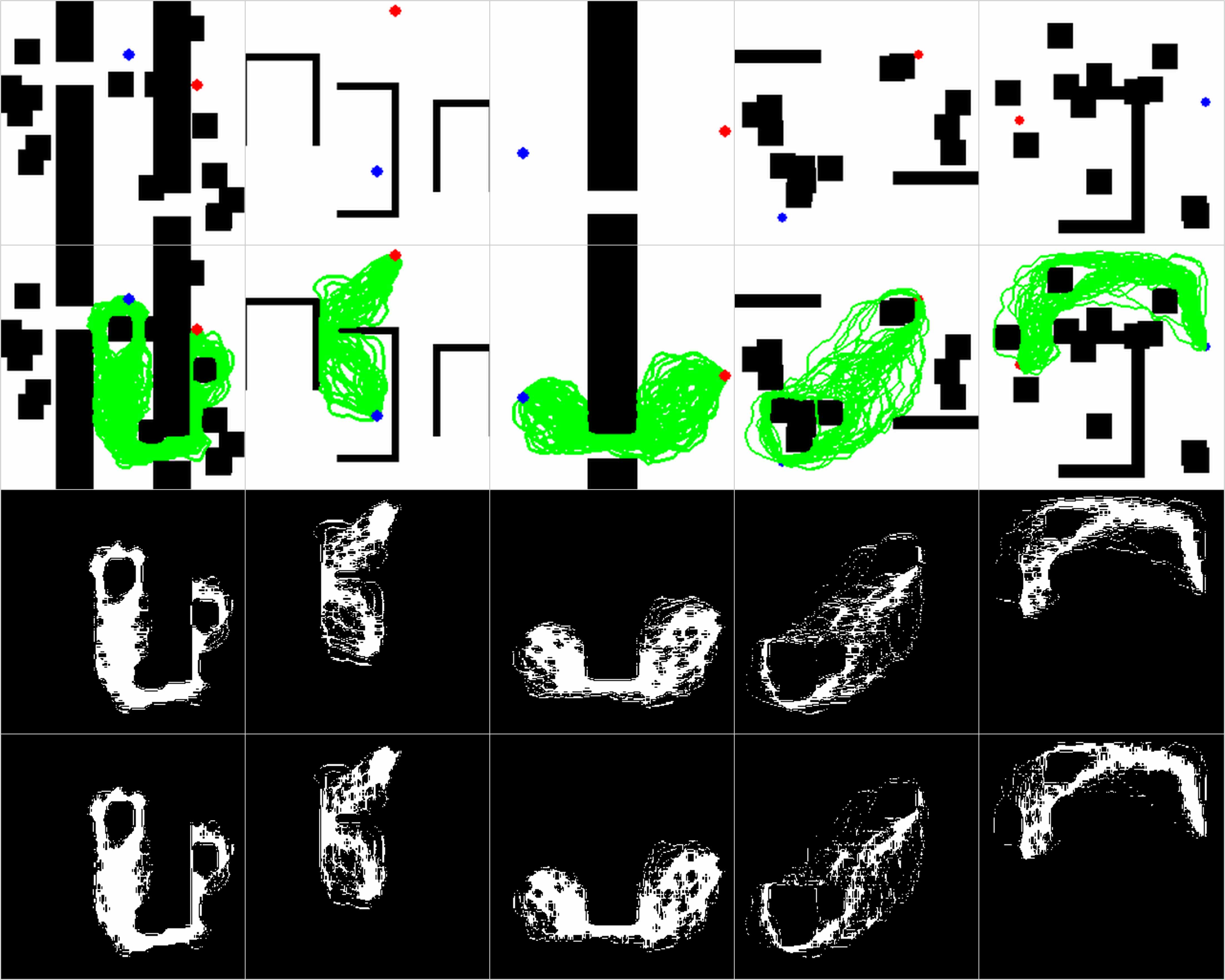}
  \caption{This figure shows 5 examples corresponding to the 5 categories of the generated dataset. 
  The first row presents 5 examples of the original maps. 
  The second row presents the feasible path sets drawn on the maps.
  The third and fourth rows present the edge labels along x-direction and y-direction, respectively.
  The green regions represent the sets of the feasible paths, and the white areas in the last two rows represent the edges lying in the promising regions.
  }
  \label{data_set}
\end{figure}

We use the same dataset as \cite{ma2021conditional}, and for clarity, we restate the generation of the dataset here.
The dataset is generated by randomly placing obstacles of different shapes on the map. 
The maps are classified into five categories according to the shape of the main obstacles. 
Fig. \ref{data_set} shows one example of each category. 
Then, we randomly sample the start and goal states on the maps, shown as red and blue dots on maps.
To obtain the corresponding ground truth where the feasible paths are more likely to locate, we run the RRT algorithm 50 times on each map.
Fifty solutions generated by RRT are drawn on the corresponding map in green, as shown in the second row of Fig. \ref{data_set}.
The green regions denote the promising regions.
Next, we generate the ground truths in the x and y directions according to the promising regions, as shown in the third and fourth rows of Fig. \ref{data_set}.
Finally, the ground truth is composed of two channels. 
One is the connectivity map in the x-direction, and the other is the connectivity map in the y-direction.
The three leftmost categories are used in training. 
The two rightmost categories are only used in testing. 
There are 12000 maps in the training set, and we select 500 of them as the test set.
In addition, we generate 500 maps of the two rightmost categories and add them to the test set.
The shape of each map is (3, 256, 256).

\subsection{Comparison of Connectivity Rates}
\label{connectivity_rate_eval}

\begin{table*}[t]
  \centering
  \caption{Comparison of connectivity rates and false negative rates.}
	\label{connect_rates_table}
  \renewcommand\arraystretch{1.3}
  \begin{tabular}{|c|c|c|c|c|}
    \hline
    \multirow{2}{*}{Methods}             & \multicolumn{2}{c|}{$Similar$}                    & \multicolumn{2}{c|}{$Dissimilar$}                   \\ 
    \cline{2-5}
    \rule{0pt}{11pt}                    ~& Connectivity Rate$\uparrow$ (\%) & False Negative Rate$\downarrow$ (\%) & Connectivity Rate$\uparrow$ (\%) & False Negative Rate$\downarrow$ (\%)   \\
    \hline
    $L_{bce} + L_{dice}$                 & 70.4                             & 24.5                                 & 76.2                             & 37.9                         \\ 
    \hline
    $CGAN$\cite{ma2021conditional}       & 91.8                             & 16.3                                 & 77.8                             & 33.0                         \\
    \hline
    $L_{bce}^{xy} + L_{dice}^{xy}$       & 98.8                             & 6.5                                  & 97.8                             & 14.5                         \\ 
    \hline
    $L_{bce}^{xy} + L_{dice}^{xy} + L_c$ & \textbf{99.4}                    & \textbf{4.9}                         & \textbf{99.8}                    & \textbf{11.9}                \\ 
    \hline
  \end{tabular}
\end{table*}

To demonstrate the performance of the proposed method, we compare it with a typical semantic segmentation method ($L_{bce} + L_{dice}$ in Table \ref{connect_rates_table}), and our previous method \cite{ma2021conditional}. 
Besides, we also conduct an ablation experiment to evaluate the efficacy of the connectivity loss $L_c$.
The connectivity rates and false negative rates obtained by all the methods involved in the comparison are summarized in Table \ref{connect_rates_table}.
The test set is divided into two subsets according to the pattern of the obstacles, as mentioned in section \ref{network_training_details}.
The map set similar to the training set contains 500 maps, named $Similar$ in Table \ref{connect_rates_table}, while the map set dissimilar to the training set also contains 500 maps, named $Dissimilar$ in Table \ref{connect_rates_table}.
The corresponding methods are indicated by the loss formats and the name of the model.
For ease of presentation, the methods in the table are marked as method 1 to method 4 from top to bottom.
Method 1 in the comparison outputs a 2-channel probabilistic map, and the first channel's values denote the likelihood to be unpromising nodes, whereas the second channel's values denote the likelihood to be promising nodes.
Method 1 is trained with BCE and Dice loss (i.e., $L_{bce}$ and $L_{dice}$), and take the node labels as the ground truth.
Method 2 is our previous method denoted as CGAN.
Method 4 is the proposed method that also outputs a 2-channel probabilistic map, and each channel represents the likelihood of the connectivity of the edges. 
Method 3 is an ablation experiment of the proposed method, where $L_c$ is not performed.

The connectivity is evaluated on the final promising region prediction results. Let $\mathcal{P}^{f1}$, $\mathcal{P}^{f2}$, $\mathcal{P}^{f3}$, and $\mathcal{P}^{f4}$ be the final prediction results of methods 1 to 4. 
The computation of $\mathcal{P}^{f1}$ is formulated below:
\begin{equation}
  \mathcal{P}^{f1}_{i,j} = \begin{cases}
    1, & if \ \hat{\mathcal{P}}^1 < \hat{\mathcal{P}}^2,\\
    0, & otherwise,   
  \end{cases}
\end{equation}
where $\{\hat{\mathcal{P}}^1, \hat{\mathcal{P}}^2 \}$ is the 2-channel output of method 1.
$\mathcal{P}^{f3}$ and $\mathcal{P}^{f4}$ are computed according to the rule defined in (\ref{final_pred_result}).
Then, we search on the final promising region prediction results (i.e., $\mathcal{P}^f$) for a feasible path from the start state to the goal state. 
If the path exists on $\mathcal{P}^f$, $\mathcal{P}^f$ is considered to be connected. Otherwise, it is disconnected. 
We calculate the connectivity rates of the predicted results of methods 1 to 4 for $Similar$ and $Dissimilar$ sets and summarize them in Table \ref{connect_rates_table}. 
Moreover, we evaluate the false-negative rates of the two test sets. 
False-negative rate reflects the degree of misprediction of the prediction result relative to ground truth.
The smaller value of the false-negative rate means the better prediction quality.
The model with the proposed method (i.e., method 3) achieves the highest connectivity rate and the lowest false-negative rate than others.
The reason is that our method takes into account the connectivity of the edges and $L_c$ guides the model to pay more attention to the promising regions.
After training, the processing speed of this model can achieve $70Hz$, which is enough to deploy it to a real-time robotic platform.

Note that our method has similar performance in both $Similar$ and $Dissimilar$ test sets. 
Nevertheless, CGAN performs in $Dissimilar$ not as well as in $Similar$.
The performance of CGAN highly depends on the training data. 
During training, it learns the transformation from the data distribution to a normal distribution and the transformation from the normal distribution to the data distribution. 
After training, it samples the normal distribution to generate the corresponding promising region. 
The model learns similar data transformations well, while the dissimilar data distribution transformations are not. 
Our model directly learns the transformation from the input map data distribution to the prediction results distribution, showing better generalization than CGAN in our task with the same training and test sets.

\section{Discussion}
\label{discussion}
Section \ref{discussion_on_diff_cr} discusses the performance of NH-RRT/RRT$^*$ with promising regions under different connectivity statuses. 
In section \ref{ros_simulation}, we implement the NH methods on ROS to demonstrate their practicability.
Section \ref{discussion_on_diff_heur} discusses the performance of NH-RRT/RRT$^*$ under different heuristic sampling biases. 
In addition, we also discuss the performance of NH-BIT$^*$ with promising regions under different connectivity statuses in section \ref{bits_discussion}.

\subsection{Discussion on Different Connectivity Statuses}
\label{discussion_on_diff_cr}

\begin{table}[t]
	\centering
	\caption{Comparison of sampling-based algorithms' performance under different connectivity statuses.
  '\textbf{d/c}' indicates \textbf{NH-RRT/RRT$^*$} with \textbf{disconnected/connected} prediction results.
  \textbf{Bold fonts} in the table indicate the best results.}
	\label{connectivity_quality_table}%
  \renewcommand{\arraystretch}{1.3}
	\setlength{\tabcolsep}{1mm}{
		\begin{tabular}{|c|c|c|c|c|c|}
			\hline
			\multicolumn{1}{|c|}{Maps} & Methods       
			& \multicolumn{1}{p{4.5em}<{\centering}|}{{Iteration Numbers$\downarrow$}} 
			& \multicolumn{1}{p{4em}<{\centering}|}{{Node Numbers$\downarrow$}} 
			& \multicolumn{1}{p{4.5em}<{\centering}|}{{Path Cost$\downarrow$}} 
			& \multicolumn{1}{p{6.5em}<{\centering}|}{{Success Rate$\uparrow$}} \\
			\hline
			\multirow{4}[1]{*}{\text{A}} 
            & \text{RRT}          & $1021 $                & $403 $               & $428.9$                  & $\textbf{100.0\%}$ \\
			& \text{d/c}  & $1211$/$\textbf{659} $ & $683$/$\textbf{364}$ & $430.9$/$\textbf{423.2}$ & $96.0\%$/$\textbf{100.0\%}$  \\
            \cline{2-6}
            & \text{RRT*}         & $1186 $                & $464 $               & $358.8$                  & $96.0\%$  \\
            & \text{d/c} & $1201$/$\textbf{721}$  & $689$/$\textbf{409}$ & $369.9$/$\textbf{344.3}$ & $94.0\%$/$\textbf{100.0\%}$    \\
			\hline

			\multirow{4}[1]{*}{\text{B}} 
            & \text{RRT}   & $873 $        & $403 $        & $502.7$         & $\textbf{100.0\%}$ \\
			& \text{d/c}   & $812 $/$\textbf{571} $ & $496 $/$\textbf{327} $ & $479.9$/$\textbf{477.3}$ & $\textbf{100.0\%}$/$\textbf{100.0\%}$ \\
            \cline{2-6}
            & \text{RRT*} & $850 $                 & $397 $                 & $438.6$                  & $\textbf{100.0\%}$ \\
            & \text{d/c}  & $816 $/$\textbf{615} $ & $502 $/$\textbf{356} $ & $\textbf{417.9}$/$418.8$ & $\textbf{100.0\%}$/$\textbf{100.0\%}$   \\
			\hline

			\multirow{4}[1]{*}{\text{C}} 
            & \text{RRT}         & $1604 $                & $901 $                & $388.1$                  & $\textbf{100.0\%}$ \\
			& \text{d/c} & $1486 $/$\textbf{541}$ & $919 $/$\textbf{358}$ & $\textbf{376.4}$/$386.6$ & $96.0\%$/$\textbf{100.0\%}$ \\
            \cline{2-6}
            & \text{RRT*}         & $1107$                & $652$                & $341.1$                  & $98.0\%$\\
            & \text{d/c} & $1190$/$\textbf{523}$ & $757$/$\textbf{353}$ & $\textbf{327.2}$/$338.0$ & $94.0\%$/$\textbf{100.0\%}$ \\
			\hline

            \multirow{4}[1]{*}{\text{D}} 
            & \text{RRT}  & $325 $                 & $216 $                 & $342.6$                  & $\textbf{100.0\%}$  \\
			& \text{d/c}  & $293 $/$\textbf{245} $ & $220 $/$\textbf{172} $ & $\textbf{330.2}$/$330.4$ & $\textbf{100.0\%}$/$\textbf{100.0\%}$ \\
            \cline{2-6}
            & \text{RRT*} & $406 $                 & $284 $                 & $289.5$                  & $\textbf{100.0\%}$ \\
            & \text{d/c}  & $306 $/$\textbf{252} $ & $234 $/$\textbf{174} $ & $\textbf{268.3}$/$275.8$ & $\textbf{100.0\%}$/$\textbf{100.0\%}$ \\
			\hline

      \multirow{4}[1]{*}{\text{E}} 
      & \text{RRT}         & $1000 $                 & $588 $                 & $453.5$                  & $\textbf{100.0\%}$ \\
			& \text{d/c} & $1394 $/$\textbf{598} $ & $569 $/$\textbf{399} $ & $417.8$/$\textbf{366.7}$ & $\textbf{100.0\%}$/$\textbf{100.0\%}$ \\
      \cline{2-6}
      & \text{RRT*}         & $966 $                  & $567 $                 & $398.1$                  & $\textbf{100.0\%}$ \\
      & \text{d/c} & $1360 $/$\textbf{627} $ & $565 $/$\textbf{420} $ & $361.4$/$\textbf{299.8}$ & $\textbf{100.0\%}$/$\textbf{100.0\%}$  \\
			\hline

      \multirow{4}[1]{*}{\text{F}} 
      & \text{RRT}  & $522 $                 & $326 $                 & $407.1$                  & $\textbf{100.0\%}$ \\
			& \text{d/c}  & $482 $/$\textbf{268} $ & $341 $/$\textbf{197} $ & $374.8$/$\textbf{368.4}$ & $\textbf{100.0\%}$/$\textbf{100.0\%}$ \\
      \cline{2-6}
      & \text{RRT*} & $546 $                 & $341 $                 & $343.9$                  & $\textbf{100.0\%}$ \\
      & \text{d/c}  & $496 $/$\textbf{290} $ & $347 $/$\textbf{212} $ & $309.6$/$\textbf{303.9}$ & $\textbf{100.0\%}$/$\textbf{100.0\%}$  \\
			\hline

      \multirow{4}[1]{*}{\text{G}} 
      & \text{RRT}         & $664 $                 & $454 $                 & $350.6$                  & $\textbf{100.0\%}$ \\
			& \text{d/c} & $531 $/$\textbf{406} $ & $383 $/$\textbf{299} $ & $303.0$/$\textbf{301.4}$ & $\textbf{100.0\%}$/$\textbf{100.0\%}$ \\
      \cline{2-6}
      & \text{RRT*}         & $704 $                & $482 $                 & $297.1$                  & $\textbf{100.0\%}$ \\
      & \text{d/c} & $611 $/$\textbf{390}$ & $442 $/$\textbf{285} $ & $\textbf{246.6}$/$256.1$ & $\textbf{100.0\%}$/$\textbf{100.0\%}$  \\
			\hline

      \multirow{4}[1]{*}{\text{H}} 
      & \text{RRT} & $268 $                 & $176 $                 & $257.6$                  & $\textbf{100.0\%}$ \\
			& \text{d/c} & $257 $/$\textbf{179} $ & $179 $/$\textbf{145} $ & $\textbf{220.9}$/$255.3$ & $\textbf{100.0\%}$/$\textbf{100.0\%}$ \\
      \cline{2-6}
      & \text{RRT*} & $282 $                 & $185 $                 & $219.9$                  & $\textbf{100.0\%}$ \\
      & \text{d/c}  & $254 $/$\textbf{183} $ & $176 $/$\textbf{147} $ & $\textbf{180.8}$/$206.0$ & $\textbf{100.0\%}$/$\textbf{100.0\%}$ \\
			\hline

	\end{tabular}}
\end{table}%

\begin{figure*}[!ht]
  \centering
  \includegraphics[width=16.0cm]{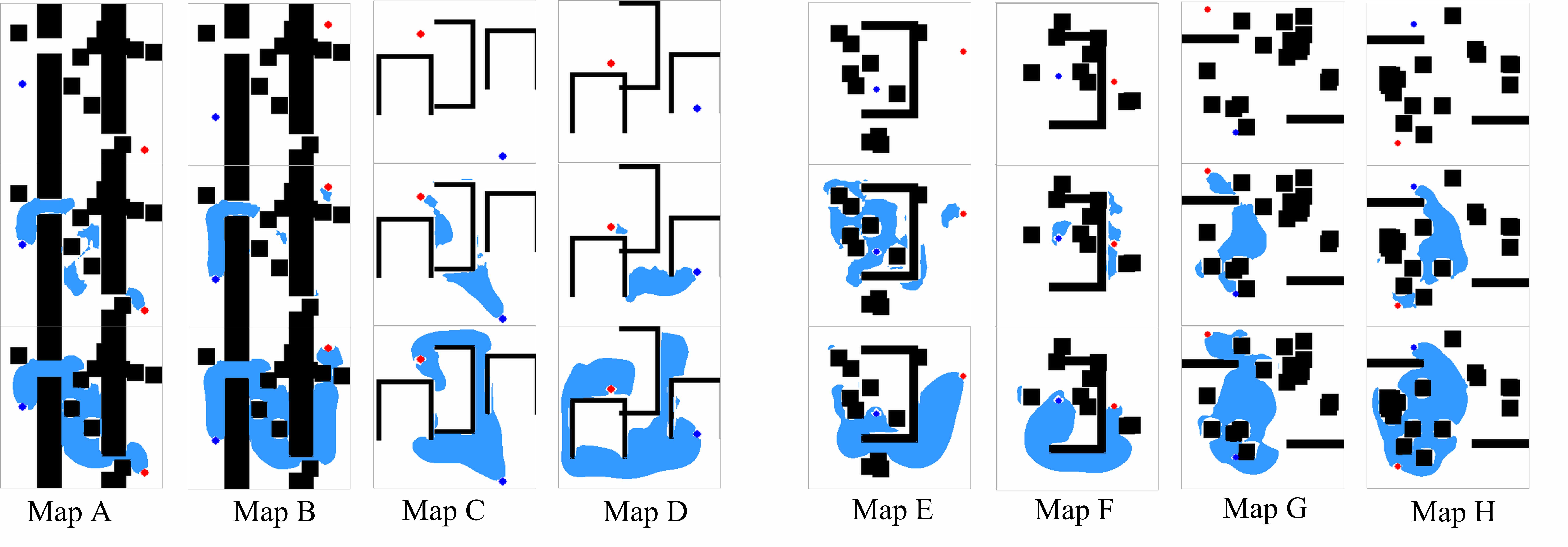}
  \caption{The first row shows the original maps and second/third shows the disconnected/connected prediction results.}
  \label{discussion_maps}
\end{figure*}

\begin{figure*}[!ht]
  \centering
  \includegraphics[width=16.0cm]{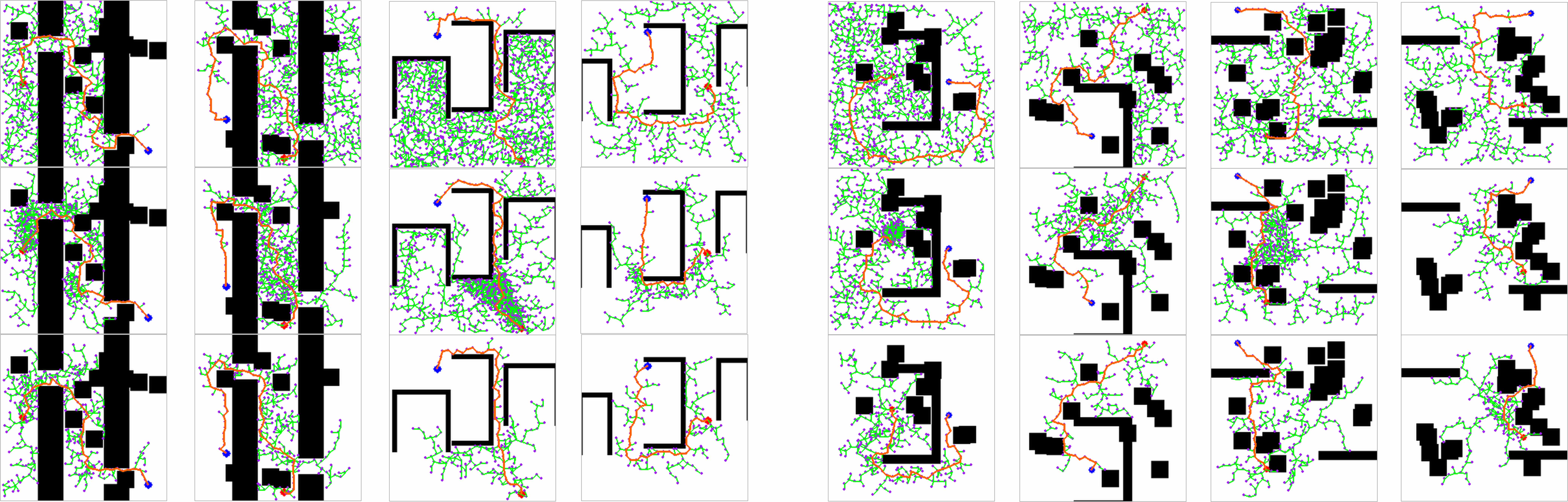}
  \caption{This figure shows successful examples of RRT in the first row and NH-RRT(d/c) in the second/third row.}
  \label{rrt_example}
\end{figure*}

\begin{figure*}[!ht]
  \centering
  \includegraphics[width=16.0cm]{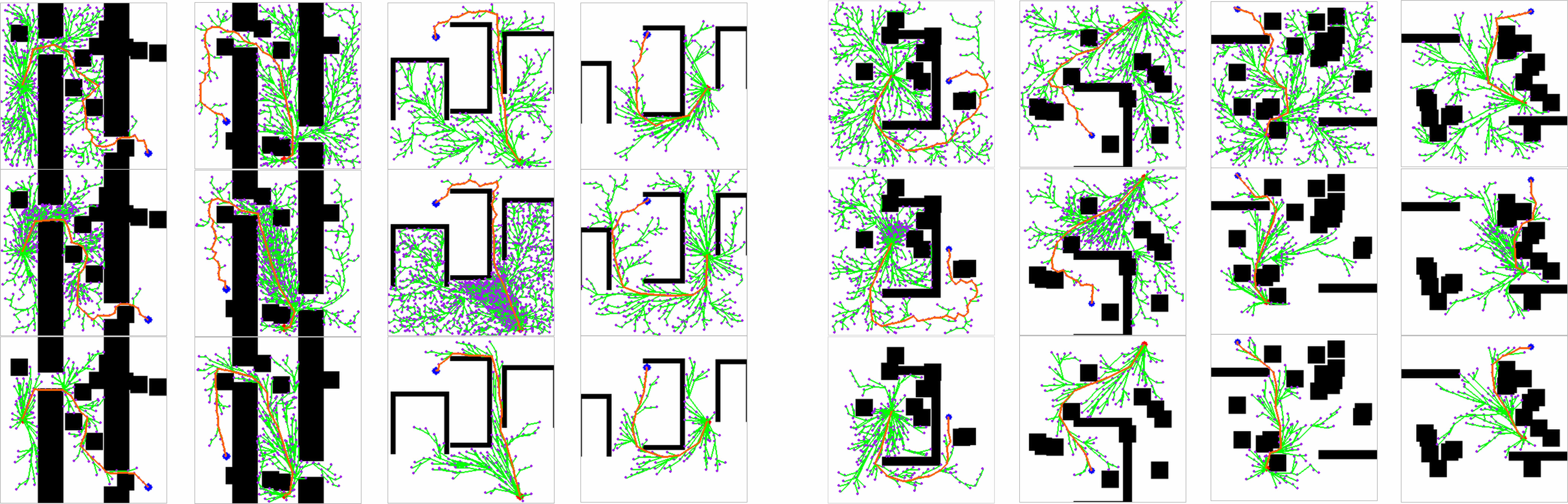}
  \caption{This figure shows successful examples of RRT$^*$ in the first row and NH-RRT$^*$(d/c) in the second/third row.}
  \label{rrtstar_example}
\end{figure*}

This section discusses the performance of NH-RRT/RRT$^*$ on maps with different connectivity statuses.
We select eight maps from four categories of the test set.
The connected results are selected from the results set with the highest connectivity rate predicted by the proposed method. 
The disconnected results are selected from the result set with the lowest connectivity rate predicted by method 1.
The visualization of the selected maps and their corresponding promising region prediction results are shown in Fig. \ref{discussion_maps}.
For ease of representation, these maps are denoted as Map A-H.
In this section, the heuristic sampling bias $h_b$ is set to $0.5$. 
We run RRT/RRT$^*$ and NH-RRT/RRT$^*$ for 50 times on each map, and the algorithms stop when they find a feasible solution or reach the maximum iteration number.
The step size of the random trees is 10.
The maximum iteration number is set to $5000$ in all the simulations.
The path cost of the initial solution, iteration numbers, node numbers of the random tree are compared in Table \ref{connectivity_quality_table}. 
Besides, we also compare the success rates under different connectivity statuses.
All the numbers in Table \ref{connectivity_quality_table} are the mean values of 50 repeated trials.

From Table \ref{connectivity_quality_table}, we find that NH-RRT/RRT$^*$ with the connected promising region reaches the best performance, especially in the iteration numbers, node numbers, and success rates.
Note that the disconnected prediction results may degrade the performance of the algorithms if a low-quality prediction result is utilized (e.g., Map E). 
More iteration and node numbers are needed to find a feasible solution with an even higher cost. 
Moreover, success rates are also reduced in some planning problems with multiple narrow passages or long narrow passages (e.g., Map A and C).
This is because the samples aggregating in a local area trap the tree from growing outwards.
The second rows of Fig. \ref{rrt_example} and Fig. \ref{rrtstar_example} give an intuition.
In some cases, NH-RRT/RRT$^*$(d) reaches similar or even a little better path cost than NH-RRT/RRT$^*$(c).
The reason is that the disconnected promising region is narrower than the connected ones and is distributed in the area where the optimal path probably lies.
By contrast, the model with our method predicts a much more accurate promising region that connects the start with the goal state, as shown in the third row of Fig. \ref{discussion_maps}.
NH-RRT/RRT$^*$ based on this prediction (i.e., NH-RRT/RRT$^*$(c)) grows towards the goal state through the promising region, which allows the random tree to find a better solution with fewer iteration numbers.
The connected predictions indicate more accurate feasible path distributions, which guarantee the best success rates in all selected maps.
The third rows of Fig. \ref{rrt_example} and Fig. \ref{rrtstar_example} give an intuition.
Without the promising region heuristic, the random tree explores the whole free space uniformly, and it may easily be trapped by the narrow passages.
With the disconnected promising region heuristic, the random tree concentrates on some local heuristic areas, which may degrade the performance of the algorithms.
However, when the predicted promising region is connected and reflects a more accurate distribution of the feasible paths, the random tree grows directly towards the goal and does not explore unnecessary areas. 
The connected promising regions significantly improve the robustness of the neural heuristic methods.

\subsection{Simulations in ROS}
\label{ros_simulation}
\begin{figure}[t]
  \centering
  \includegraphics[width=8.8cm]{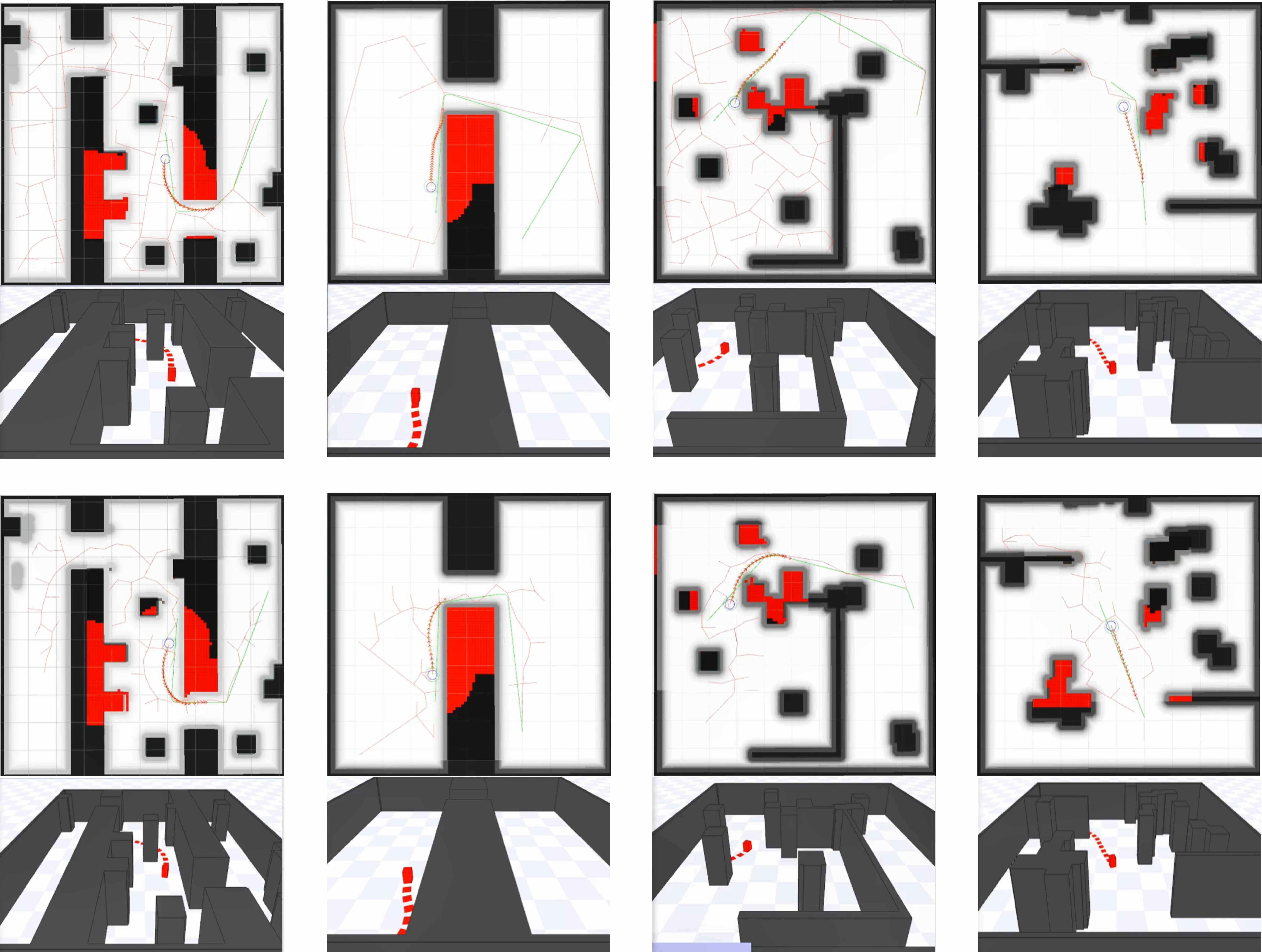}
  \caption{The first and second rows are navigation processes without heuristic sampling (RRT), and the third and fourth rows are with heuristic sampling (NH-RRT). Both 2D and 3D views are shown for each example. In 2D views, the circles denote the robot, and the red edges are the edges for the random trees, and the green curves denote the global plans, and the red arrows are the local plan generated by TEB local planner. In 3D views, the red cuboids represent the robot, and the following red squares are its footprints.
  }
  \label{ros_rrt}
\end{figure}

\begin{figure}[ht]
  \centering
  \includegraphics[width=8.8cm]{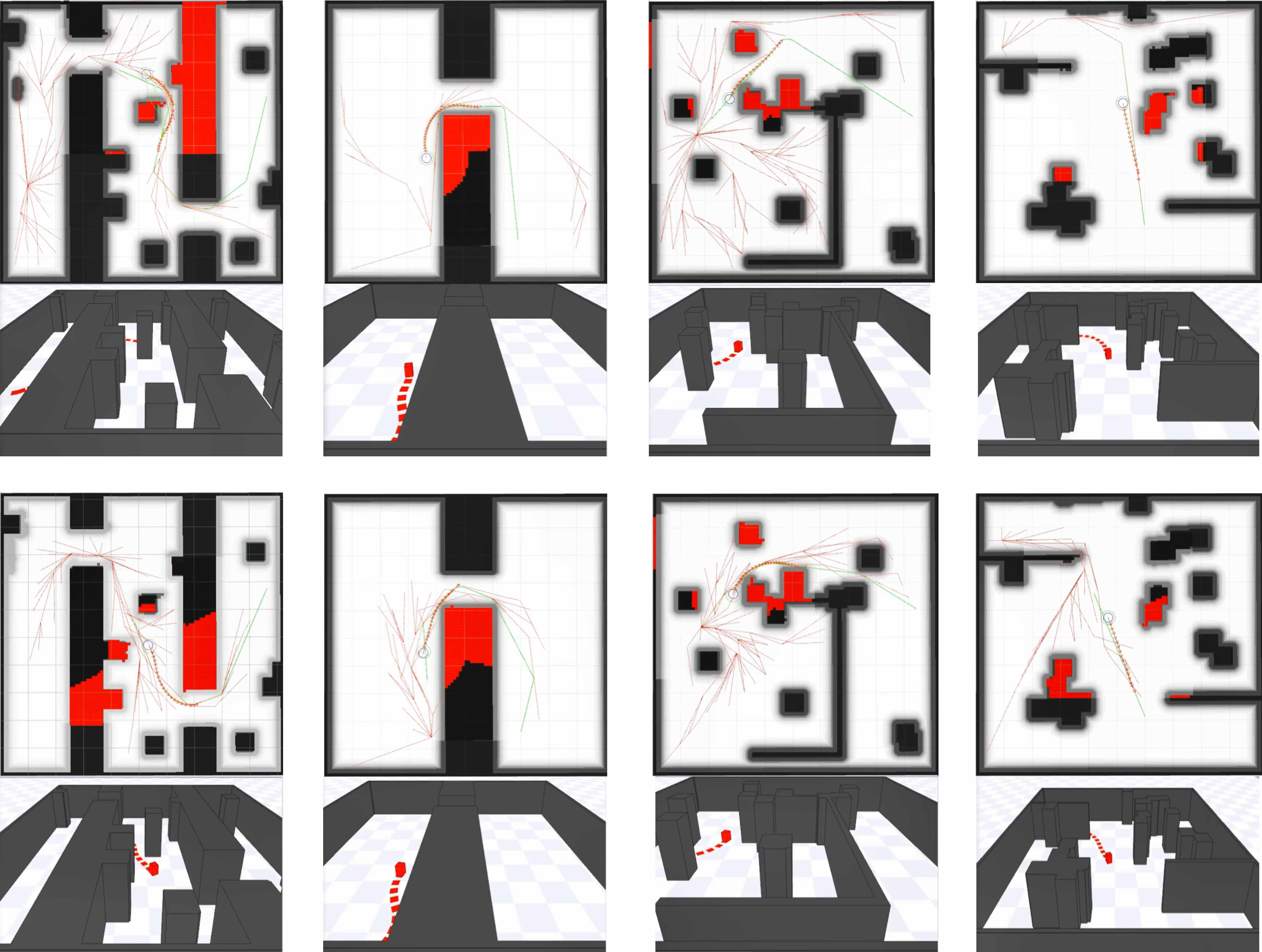}
  \caption{The first and second rows are navigation processes without heuristic sampling (RRT$^*$), and the third and fourth rows are with heuristic sampling (NH-RRT$^*$). Both 2D and 3D views are shown for each example.
  }
  \label{ros_rrtstar}
\end{figure}

\begin{table*}[ht]
	\centering
	\caption{Average Planning Time (PT) and Navigation Time (NT) on four maps in ROS simulation.}
	\label{ros_results}
	\begin{tabular}{ |c|c|c|c|c|c|c|c|c|} 
		\hline
		\multirow{2}{*}{Methods}& \multicolumn{2}{|c|}{Map \uppercase\expandafter{\romannumeral1}} & \multicolumn{2}{|c|}{Map \uppercase\expandafter{\romannumeral2}} & \multicolumn{2}{|c|}{Map \uppercase\expandafter{\romannumeral3}} & \multicolumn{2}{|c|}{Map \uppercase\expandafter{\romannumeral4}} \\
		\cline{2-9}
		                      ~&PT(s)$\downarrow$ & NT(m)$\downarrow$ & PT(s)$\downarrow$ & NT(m)$\downarrow$ & PT(s)$\downarrow$ & NT(m)$\downarrow$ & PT(s)$\downarrow$ & NT(m)$\downarrow$\\
		\hline
		RRT  & 45.139 & 97.223 & 8.811 & 54.148 & 21.745 & 53.845 & 0.916 & 32.488\\
		\hline
		NH-RRT & \textbf{0.52} & \textbf{51.878} & \textbf{1.134} & \textbf{45.955} & \textbf{0.156} & \textbf{29.756} & \textbf{0.198} & \textbf{30.793}\\
		\hline
		RRT$^*$  & 39.367 & 90.147 & 14.115 & 59.229 & 14.73 & 47.582 & 1.531 & 32.902\\
		\hline
		NH-RRT$^*$ & \textbf{1.534} & \textbf{50.492} & \textbf{1.271} & \textbf{46.068} & \textbf{0.586} & \textbf{30.457} & \textbf{0.477} & \textbf{31.01}\\
		\hline
	\end{tabular}
\end{table*}

In addition, we conduct simulation experiments based on ROS to demonstrate the practicability of our method. 
In the simulations of this section, we select four maps from the test set, which are shown in Fig. \ref{ros_rrt} and Fig. \ref{ros_rrtstar}.
For convenience, we name these four maps as Map \uppercase\expandafter{\romannumeral1}, Map \uppercase\expandafter{\romannumeral2}, Map \uppercase\expandafter{\romannumeral3} and Map \uppercase\expandafter{\romannumeral4} from left to right in Fig. \ref{ros_rrt}.
Map \uppercase\expandafter{\romannumeral1} and Map \uppercase\expandafter{\romannumeral2} are from $Similar$, and Map \uppercase\expandafter{\romannumeral3} and Map \uppercase\expandafter{\romannumeral4} are from $Dissimilar$.
The resolution of these maps is set to $0.04m/pixel$.
We use the ROS map server to define the maps.
A differential-driven mobile robot with the size of $0.25m \times 0.25m \times 0.4m$ is used in the navigation tasks, which is also installed with a laser scanner for localization.
We use the ROS Stage platform to define the robot with sensors.
Each map corresponds to a navigation task, in which the robot needs to navigate from a start state to a goal state.
In the simulations,  RRT/RRT$^*$ or NH-RRT/RRT$^*$ serves as the global planner, and Timed-Elastic-Bands (TEB) \cite{teb_planner} serves as the local planner.
All the algorithm runs until the solution is found, and the step size of the random trees are set to $0.5m$.
The linear velocity range of the robot is $[-0.2, 0.4] m/s$, and the angular velocity range of the robot is $[-0.3, 0.3] rad/s$.
Note that we optimize the path to get the final global plan via  Lazy States Contraction (LSC) \cite{hauser2010fast}, and all the algorithms are implemented in C++.
We compare the performance of RRT/RRT$^*$ with NH-RRT/RRT$^*$ in terms of average Planning Time (PT) and average Navigation Time (NT). 
PT indicates the total planning time consumption during a navigation task, while NT indicates the time taken by the robot from receiving the goal state to reaching the goal state. 
For NH-RRT/RRT$^*$, we use the heuristic generated by our method and set $h_b$ to $0.5$. 

Fig. \ref{ros_rrt} and Fig. \ref{ros_rrtstar} show the scenes where the robot is navigating on the maps, in 2D and 3D views. 
Each navigation task is solved 30 times by each method, and we record the average PT and NT corresponding to each method in Table \ref{ros_results}. 
We find that NH-RRT/RRT$^*$ consumes much less average PT and NT than RRT/RRT$^*$ for all the navigation tasks in the simulations.
The improvement is significant, especially in the hard task in Map \uppercase\expandafter{\romannumeral1}.
The reason is that many unnecessary samples are not involved in the planning procedure with the neural heuristic, and many samples are drawn from the promising region.
Therefore, the random trees can extend more aggressively towards the goal state.

\subsection{Discussion on Different Heuristic Sampling Bias}
\label{discussion_on_diff_heur}

\begin{table}[t]
	\centering
	\caption{Comparison of the algorithms' performance under different heuristic sampling biases.
  '\textbf{d/c  \boldmath$h_b$}' represents that \textbf{NH-RRT/RRT*} with the \textbf{disconnected/connected} prediction result run when heuristic sampling bias equals  \boldmath$h_b$. 
  \textbf{Bold fonts} in the table indicate the best results.}
	\label{sampling_bias_table}%
  \renewcommand{\arraystretch}{1.3}
	\setlength{\tabcolsep}{1mm}{
		\begin{tabular}{|c|c|c|c|c|c|}
			\hline
			\multicolumn{1}{|c|}{Maps} & \multicolumn{1}{p{3.2em}<{\centering}|}{Methods}       
			& \multicolumn{1}{p{4.5em}<{\centering}|}{{Iteration Numbers$\downarrow$}} 
			& \multicolumn{1}{p{4em}<{\centering}|}{{Node Numbers$\downarrow$}} 
			& \multicolumn{1}{p{4.5em}<{\centering}|}{{Path Cost$\downarrow$}} 
			& \multicolumn{1}{p{6.5em}<{\centering}|}{{Success Rate$\uparrow$}} \\
			\hline
			\multirow{8}[1]{*}{\text{A}}
      & \text{RRT}      & $1021 $                 & $403 $                 & $428.9$                  & $\textbf{100.0\%}$ \\
			& \text{d/c 0.3}  & $1197 $/$\textbf{716} $ & $607 $/$\textbf{359} $ & $426.7$/$\textbf{417.6}$ & $98.0\%$/$\textbf{100.0\%}$ \\
      & \text{d/c 0.6}  & $1208 $/$\textbf{596} $ & $743 $/$\textbf{349} $ & $421.2$/$\textbf{415.0}$ & $88.0\%$/$\textbf{100.0\%}$ \\
      & \text{d/c 0.9}  & $1284 $/$\textbf{416} $ & $893 $/$\textbf{272} $ & $411.3$/$\textbf{406.5}$ & $80.0\%$/$\textbf{100.0\%}$  \\
      \cline{2-6}
      & \text{RRT*}    & $1186 $ & $464 $ & $358.8$ & $96.0\%$ \\
      & \text{d/c 0.3} & $887 $/$\textbf{800} $  & $456 $/$\textbf{398} $ & $363.7$/$\textbf{347.8}$ & $92.0\%$/$\textbf{100.0\%}$ \\
      & \text{d/c 0.6} & $1035 $/$\textbf{695} $ & $639 $/$\textbf{409} $ & $360.7$/$\textbf{346.0}$ & $90.0\%$/$\textbf{100.0\%}$  \\
      & \text{d/c 0.9} & $1361 $/$\textbf{654} $ & $921 $/$\textbf{433} $ & $360.5$/$\textbf{339.8}$ & $76.0\%$/$\textbf{100.0\%}$ \\
			\hline
			
      \multirow{8}[1]{*}{\text{C}} 
      & \text{RRT}     & $1604 $                 & $901 $                  & $388.1$                   & $\textbf{100.0\%}$ \\
			& \text{d/c 0.3} & $1570 $/$\textbf{674} $ & $941 $/$\textbf{432} $  & $\textbf{383.5}$/$384.6$  & $98.0\%$/$\textbf{100.0\%}$ \\
      & \text{d/c 0.6} & $1267 $/$\textbf{451} $ & $807 $/$\textbf{308} $  & $\textbf{367.9}$/$381.7$  & $88.0\%$/$\textbf{100.0\%}$ \\
      & \text{d/c 0.9} & $2427 $/$\textbf{355} $ & $1580 $/$\textbf{256} $ & $\textbf{371.1}$/$379.3$  & $76.0\%$/$\textbf{100.0\%}$ \\
      \cline{2-6}
      & \text{RRT*}     & $1107 $                 & $652 $                  & $341.1$                  & $98.0\%$ \\
      & \text{d/c 0.3}  & $1452 $/$\textbf{711} $ & $873 $/$\textbf{452} $  & $\textbf{327.0}$/$333.3$ & $96.0\%$/$\textbf{100.0\%}$ \\
      & \text{d/c 0.6}  & $1242 $/$\textbf{451} $ & $802 $/$\textbf{313} $  & $\textbf{331.5}$/$337.1$ & $88.0\%$/$\textbf{100.0\%}$ \\
      & \text{d/c 0.9}  & $2064 $/$\textbf{388} $ & $1339 $/$\textbf{278} $ & $\textbf{323.8}$/$332.9$ & $84.0\%$/$\textbf{100.0\%}$  \\
			\hline

      \multirow{8}[1]{*}{\text{E}} 
      & \text{RRT}     & $1000 $                 & $588 $                 & $453.5$                  & $\textbf{100.0\%}$ \\
			& \text{d/c 0.3} & $1082 $/$\textbf{841} $ & $\textbf{512} $/$527 $ & $410.4$/$\textbf{401.3}$ & $\textbf{100.0\%}$/$\textbf{100.0\%}$ \\
      & \text{d/c 0.6} & $1680 $/$\textbf{521} $ & $645 $/$\textbf{359} $ & $408.5$/$\textbf{350.7}$ & $\textbf{100.0\%}$/$\textbf{100.0\%}$ \\
      & \text{d/c 0.9} & $4056 $/$\textbf{426} $ & $1202$/$\textbf{317}$  & $377.9$/$\textbf{352.7}$ & $22.0\%$/$\textbf{100.0\%}$ \\
      \cline{2-6}
      & \text{RRT*}    & $966$                  & $567 $                 & $398.1$                  & $\textbf{100.0\%}$ \\
      & \text{d/c 0.3} & $1129$/$\textbf{813}$  & $531$/$\textbf{514}$   & $372.6$/$\textbf{326.5}$ & $\textbf{100.0\%}$/$\textbf{100.0\%}$  \\
      & \text{d/c 0.6} & $1793$/$\textbf{585}$  & $685$/$\textbf{399}$   & $375.4$/$\textbf{290.0}$ & $\textbf{100.0\%}$/$\textbf{100.0\%}$  \\
      & \text{d/c 0.9} & $3812$/$\textbf{436}$  & $1144$/$\textbf{326}$  & $351.9$/$\textbf{282.3}$ & $16.0\%$/$\textbf{100.0\%}$  \\
			\hline

      \multirow{8}[1]{*}{\text{G}} 
      & \text{RRT}     & $664 $                  & $454 $                 & $350.6$                  & $\textbf{100.0\%}$ \\
			& \text{d/c 0.3} & $578 $/$\textbf{490} $  & $412 $/$\textbf{353} $ & $\textbf{306.3}$/$320.0$ & $\textbf{100.0\%}$/$\textbf{100.0\%}$ \\
      & \text{d/c 0.6} & $655 $/$\textbf{416} $  & $468 $/$\textbf{306} $ & $\textbf{301.2}$/$303.7$ & $\textbf{100.0\%}$/$\textbf{100.0\%}$ \\
      & \text{d/c 0.9} & $1031 $/$\textbf{341} $ & $773 $/$\textbf{256} $ & $\textbf{290.0}$/$293.5$ & $96.0\%$/$\textbf{100.0\%}$ \\
      \cline{2-6}
      & \text{RRT*}     & $704 $ & $482 $ & $297.1$ & $\textbf{100.0\%}$ \\
      & \text{d/c 0.3}  & $\textbf{511} $/$512 $  & $\textbf{361} $/$370 $  & $\textbf{266.2}$/$268.7$ & $\textbf{100.0\%}$/$\textbf{100.0\%}$ \\
      & \text{d/c 0.6}  & $625 $/$\textbf{391} $  & $448 $/$\textbf{291} $  & $252.3$/$\textbf{251.1}$ & $\textbf{100.0\%}$/$\textbf{100.0\%}$ \\
      & \text{d/c 0.9}  & $1097 $/$\textbf{352} $ & $819 $/$\textbf{266} $  & $\textbf{242.7}$/$243.2$ & $98.0\%$/$\textbf{100.0\%}$ \\
			\hline
	\end{tabular}}
  \vspace{-1mm}
\end{table}%

In this section, we discuss the effects of different heuristic sampling biases ($h_b$) on the performance of NH-RRT/RRT$^*$(d/c).
$h_b$ is set to $0.0$, $0.3$, $0.6$, and $0.9$. 
We also run NH-RRT/RRT$^*$ for 50 times under each $h_b$ setting. 
The iteration numbers, node numbers of the random tree, path cost of the initial solution, and success rate are considered in the comparison.
Table \ref{sampling_bias_table} lists all the results on Map A, C, E, and G. 
All the numbers in Table \ref{sampling_bias_table} are the mean values of 50 repeated trials.
When $h_b=0.0$, NH-RRT/RRT$^*$ degenerates to the original RRT/RRT$^*$. 
Thus, we use the name of the original algorithms to denote $h_b=0.0$ in Table \ref{sampling_bias_table}.

According to Table \ref{sampling_bias_table}, in most cases, the performance of NH-RRT/RRT$^*$ with the connected prediction results (i.e., NH-RRT/RRT$^*$(c)) improves as $h_b$ increases.
In the four maps, NH-RRT/RRT$^*$(c) with all $h_b$ settings achieves the highest success rate while taking the minor iteration numbers.
In most cases, NH-RRT/RRT$^*$(c) also generates the fewest node numbers and reaches the lowest path cost.
However, in some cases, the performance of NH-RRT/RRT$^*$(d) may decrease as $h_b$ increases.
The reason is that the disconnected promising region predictions hinder the exploration of the random trees when $h_b$ is large.
On the contrary, the connected promising region predictions contain some feasible solutions, which prevent the random tree from being trapped in a local area.
In Map C and Map G, NH-RRT/RRT$^*$(d) reaches similar or even a little better path cost than NH-RRT/RRT$^*$(c). 
The reason is that the disconnected promising region is narrower than the connected ones and is distributed in the area where the optimal path probably lies.
Besides, when $h_b$ is set to a large value and the promising region prediction is connected, the random tree can find a feasible solution even more quickly.
A good initial solution is beneficial for algorithms like RRT$^*$ to converge to the optimal solution.
On the other hand, a connected prediction result can also reduce the NH algorithms' dependence on the heuristic sampling bias.

\subsection{Discussion on Neural Heuristic BIT$^*$}
\label{bits_discussion}
\begin{figure*}[!ht]
  \centering
  \includegraphics[width=16.0cm]{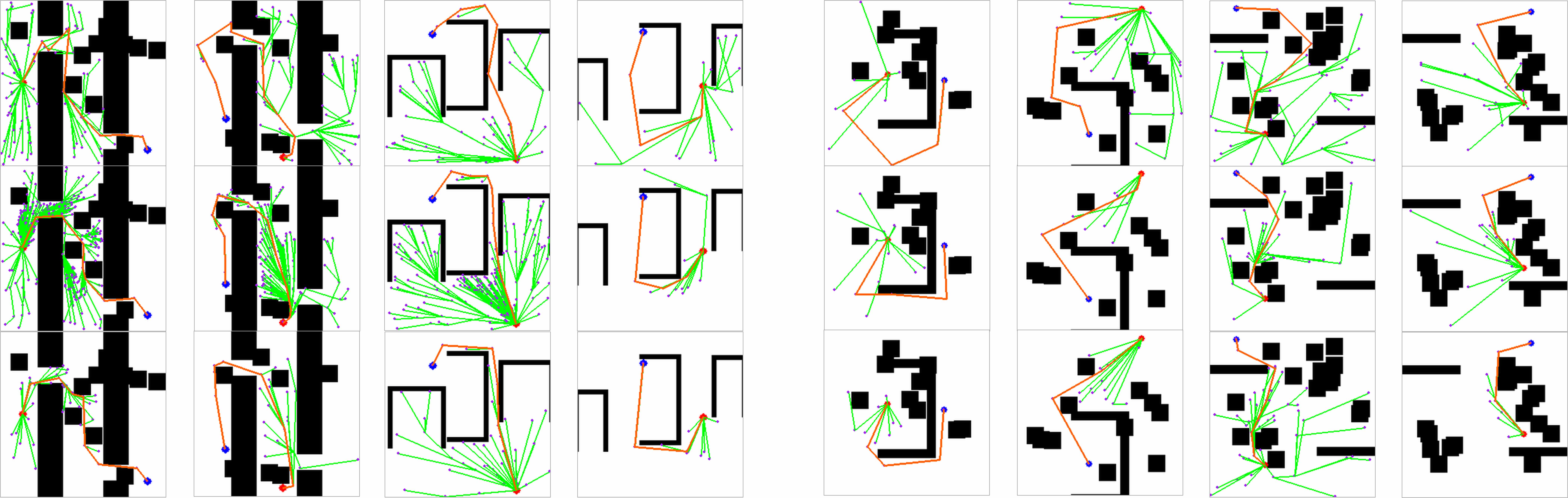}
  \caption{
  This figure shows successful examples of BIT$^*$ in the first row and NH-BIT$^*$(d)/(c) in the second/third row.
  }
  \label{bitstar_example}
\end{figure*}

\begin{table}[ht]
	\centering
	\caption{Comparison of sampling-based algorithms' performance under different connectivity statuses.
  '\textbf{d/c}' indicates \textbf{NH-BIT*} with \textbf{disconnected/connected} prediction results.
  \textbf{Bold fonts} in the table indicate the best results.}
	\label{bitstar_connectivity_quality_table}%
  \renewcommand{\arraystretch}{1.3}
	\setlength{\tabcolsep}{1mm}{
		\begin{tabular}{|c|c|c|c|c|c|}
			\hline
			\multicolumn{1}{|c|}{Maps} & Methods       
			& \multicolumn{1}{p{4.5em}<{\centering}|}{{Iteration Numbers$\downarrow$}} 
			& \multicolumn{1}{p{4em}<{\centering}|}{{Node Numbers$\downarrow$}} 
			& \multicolumn{1}{p{4.5em}<{\centering}|}{{Path Cost$\downarrow$}} 
			& \multicolumn{1}{p{6.5em}<{\centering}|}{{Success Rate$\uparrow$}} \\
			\hline
			\multirow{2}[1]{*}{\text{A}} 
            & \text{BIT*}          & $1746 $ & $155 $ & $378.5$ & $94.0\%$  \\
			& \text{d/c 0.5}  & $8660 $/$\textbf{593} $  & $163 $/$\textbf{63}$ & $\textbf{363.1}$/$373.0$ & $90.0\%$/$\textbf{100.0\%}$   \\
			\hline

            \multirow{2}[1]{*}{\text{B}} 
            & \text{BIT*}   & $\textbf{334} $ & $67 $ & $445.1$ & $\textbf{100.0\%}$\\
			& \text{d/c 0.5}    & $901 $/$426 $ & $111 $/$\textbf{58} $ & $\textbf{422.4}$/ $424.8$ & $\textbf{100.0\%}$/$\textbf{100.0\%}$    \\
			\hline

            \multirow{2}[1]{*}{\text{C}}
            & \text{BIT*}    & $1135 $ & $84 $ & $350.5$ & $\textbf{100.0\%}$  \\
			& \text{d/c 0.5}& $11479 $/$\textbf{254} $& $175 $/$\textbf{42}$ & $\textbf{344.7}$/$346.7$ & $96.0\%$/$\textbf{100.0\%}$  \\
			\hline

            \multirow{2}[1]{*}{\text{D}}
            & \text{BIT*}  & $\textbf{78} $ & $16 $ & $302.5$ & $\textbf{100.0\%}$   \\
			& \text{d/c 0.5}  & $105 $/$100 $  & $17 $/$\textbf{14}$ & $\textbf{278.6}$/$287.2$ & $\textbf{100.0\%}$/$\textbf{100.0\%}$ \\
			\hline

            \multirow{2}[1]{*}{\text{E}}
            & \text{BIT*}  & $\textbf{91}$ & $\textbf{21}$ & $352.9$ & $\textbf{100.0\%}$  \\
        	& \text{d/c 0.5}  & $410 $/$155 $ & $27 $/ $24 $ & $370.4$/$\textbf{298.2}$  & $\textbf{100.0\%}$/$\textbf{100.0\%}$  \\
			\hline
			
			\multirow{2}[1]{*}{\text{F}}
            & \text{BIT*}    & $\textbf{87}$ & $23 $ & $331.9$ & $\textbf{100.0\%}$ \\
    		& \text{d/c 0.5} & $207 $/$104 $ & $29 $/ $\textbf{20}$ & $309.9$/$\textbf{299.7}$  & $\textbf{100.0\%}$/$\textbf{100.0\%}$ \\
			\hline

            \multirow{2}[1]{*}{\text{G}}
            & \text{BIT*}   & $\textbf{161} $ & $37 $ & $353.4$ & $\textbf{100.0\%}$ \\
			& \text{d/c 0.5} & $513 $/$176 $ & $52 $/$\textbf{28} $ & $\textbf{296.3}$/$\textbf{296.3}$ & $\textbf{100.0\%}$/$\textbf{100.0\%}$  \\
			\hline

            \multirow{2}[1]{*}{\text{H}} 
            & \text{BIT*} & $60 $ & $\textbf{14} $ & $233.7$ & $\textbf{100.0\%}$ \\
			& \text{d/c 0.5} & $120 $/$\textbf{51} $ & $18 $/$\textbf{14} $ & $215.2$/ $\textbf{210.2}$  & $\textbf{100.0\%}$/$\textbf{100.0\%}$ \\
			\hline
	\end{tabular}}
\end{table}%

BIT$^*$ is one of the state-of-the-art sampling-based algorithms.
In this section, we discuss the performance of BIT$^*$ with our method (NH-BIT$*$) to show our method's capability to combine with the state-of-the-art sampling-based algorithm. 
Since BIT$^*$ is an informed tree-based method that samples in an informed ellipsoid after finding the initial solution, the heuristic sampling of our method may disturb its original informed sampling.
Therefore, we only consider the stage where BIT$^*$ searches for the initial solution.
The sampling process of BIT$^*$ is similar to RRT/RRT$^*$ except that it samples a batch of samples once a time. Thus, the modification of the sampling strategy in NH-BIT$^*$ is the same as that of NH-RRT/RRT$^*$.
The batch size of BIT$^*$/NH-BIT$^*$ is set to $30$, and we run BIT$^*$/NH-BIT$^*$ 50 times on each map.
The $h_b$ of NH-BIT$^*$ is set to $0.5$. 
The maximum run time of BIT$^*$/NH-BIT$^*$ is limited to $10s$.
The maximum sample number of BIT$^*$/NH-BIT$^*$ is limited to $1000$. 
Fig. \ref{bitstar_example} presents some examples of BIT$^*$ and NH-BIT$^*$ with the disconnected or connected promising region.
The success rate, average iteration numbers, node numbers, and path cost of 50 repeated trials on each map are presented in Table \ref{bitstar_connectivity_quality_table}.
It is observed that the average path costs achieved by NH-BIT$^*$ are smaller than those achieved by BIT$^*$, and NH-BIT$^*$(c) always has the highest success rate. 
However, the average iteration numbers may increase with heuristic sampling.
The reason is that some of the samples in one batch are gathered in the promising region.
In other words, the extension of the tree with heuristic sampling will be slower than that without NH, but NH reduces the cost of the initial solution. 
In all test cases, NH-BIT$^*$(c) increases the iteration numbers to a smaller extent than NH-BIT$^*$(d), and NH-BIT$^*$(c) can reduce the iteration numbers in some complex problems such as Map A and C.
Therefore, NH-BIT$^*$(c) can improve the performance of BIT* more stably and has the ability to deal with more complex problems.

\section{Conclusion and Future Work}
\label{conclusions}
We present a novel method to enhance the connectivity of the promising region predicted by neural networks.
The simulation results demonstrate that our method improves the connectivity rates of the promising region predictions.
Moreover, we discuss the effects of the connectivity status of the prediction results on the neural heuristic sampling-based algorithms.
The results show that the connected promising region predictions ensure the robustness of the algorithms and reduce the dependence on heuristic sampling bias. 
Therefore, ensuring the connectivity of promising regions is necessary for applying neural heuristics in sampling-based path planning algorithms.
We also implement ROS simulations that demonstrate the practicability of NH sampling-based methods.

In future work, we hope to extend our method to 3D environments.
The extension can be achieved by substituting 2D convolutional layers with 3D convolutional layers and adding one more direction in map representation and loss computations.

\ifCLASSOPTIONcaptionsoff
\newpage
\fi

\bibliographystyle{IEEEtran}
\bibliography{reference}

\begin{thebibliography}{10}
\providecommand{\url}[1]{#1}
\csname url@samestyle\endcsname
\providecommand{\newblock}{\relax}
\providecommand{\bibinfo}[2]{#2}
\providecommand{\BIBentrySTDinterwordspacing}{\spaceskip=0pt\relax}
\providecommand{\BIBentryALTinterwordstretchfactor}{4}
\providecommand{\BIBentryALTinterwordspacing}{\spaceskip=\fontdimen2\font plus
\BIBentryALTinterwordstretchfactor\fontdimen3\font minus
  \fontdimen4\font\relax}
\providecommand{\BIBforeignlanguage}[2]{{%
\expandafter\ifx\csname l@#1\endcsname\relax
\typeout{** WARNING: IEEEtran.bst: No hyphenation pattern has been}%
\typeout{** loaded for the language `#1'. Using the pattern for}%
\typeout{** the default language instead.}%
\else
\language=\csname l@#1\endcsname
\fi
#2}}
\providecommand{\BIBdecl}{\relax}
\BIBdecl

\bibitem{dijkstra1959note}
E.~W. Dijkstra \emph{et~al.}, ``A note on two problems in connexion with
  graphs,'' \emph{Numerische mathematik}, vol.~1, no.~1, pp. 269--271, 1959.

\bibitem{hart1968formal}
P.~E. Hart, N.~J. Nilsson, and B.~Raphael, ``A formal basis for the heuristic
  determination of minimum cost paths,'' \emph{IEEE transactions on Systems
  Science and Cybernetics}, vol.~4, no.~2, pp. 100--107, 1968.

\bibitem{khatib1986real}
O.~Khatib, ``Real-time obstacle avoidance for manipulators and mobile robots,''
  in \emph{Autonomous robot vehicles}.\hskip 1em plus 0.5em minus 0.4em\relax
  Springer, 1986, pp. 396--404.

\bibitem{ferguson2004focussed}
D.~Ferguson and A.~Stentz, ``Focussed processing of mdps for path planning,''
  in \emph{16th IEEE International Conference on Tools with Artificial
  Intelligence}.\hskip 1em plus 0.5em minus 0.4em\relax IEEE, 2004, pp.
  310--317.

\bibitem{bakker2005hierarchical}
B.~Bakker, Z.~Zivkovic, and B.~Krose, ``Hierarchical dynamic programming for
  robot path planning,'' in \emph{2005 IEEE/RSJ International Conference on
  Intelligent Robots and Systems}.\hskip 1em plus 0.5em minus 0.4em\relax IEEE,
  2005, pp. 2756--2761.

\bibitem{lavalle2001randomized}
S.~M. LaValle and J.~J. Kuffner~Jr, ``Randomized kinodynamic planning,''
  \emph{The international journal of robotics research}, vol.~20, no.~5, pp.
  378--400, 2001.

\bibitem{kavraki1996probabilistic}
L.~E. Kavraki, P.~Svestka, J.-C. Latombe, and M.~H. Overmars, ``Probabilistic
  roadmaps for path planning in high-dimensional configuration spaces,''
  \emph{IEEE transactions on Robotics and Automation}, vol.~12, no.~4, pp.
  566--580, 1996.

\bibitem{karaman2010incremental}
S.~Karaman and E.~Frazzoli, ``Incremental sampling-based algorithms for optimal
  motion planning,'' \emph{Robotics Science and Systems VI}, vol. 104, no.~2,
  2010.

\bibitem{gammell2014informed}
J.~D. Gammell, S.~S. Srinivasa, and T.~D. Barfoot, ``Informed rrt*: Optimal
  sampling-based path planning focused via direct sampling of an admissible
  ellipsoidal heuristic,'' in \emph{2014 IEEE/RSJ International Conference on
  Intelligent Robots and Systems}.\hskip 1em plus 0.5em minus 0.4em\relax IEEE,
  2014, pp. 2997--3004.

\bibitem{xu2021learning}
S.~Xu, J.~Liu, C.~Yang, X.~Wu, and T.~Xu, ``A learning-based stable servo
  control strategy using broad learning system applied for microrobotic
  control,'' \emph{IEEE Transactions on Cybernetics}, 2021.

\bibitem{wang2020neural}
J.~Wang, W.~Chi, C.~Li, C.~Wang, and M.~Q.-H. Meng, ``Neural rrt*:
  Learning-based optimal path planning,'' \emph{IEEE Transactions on Automation
  Science and Engineering}, vol.~17, no.~4, pp. 1748--1758, 2020.

\bibitem{ma2021conditional}
N.~Ma, J.~Wang, J.~Liu, and M.~Q.-H. Meng, ``Conditional generative adversarial
  networks for optimal path planning,'' \emph{IEEE Transactions on Cognitive
  and Developmental Systems}, 2021.

\bibitem{zhang2021generative}
T.~Zhang, J.~Wang, and M.~Q.-H. Meng, ``Generative adversarial network based
  heuristics for sampling-based path planning,'' \emph{IEEE/CAA Journal of
  Automatica Sinica}, vol.~9, no.~1, pp. 64--74, 2021.

\bibitem{janson2015fast}
L.~Janson, E.~Schmerling, A.~Clark, and M.~Pavone, ``Fast marching tree: A fast
  marching sampling-based method for optimal motion planning in many
  dimensions,'' \emph{The International journal of robotics research}, vol.~34,
  no.~7, pp. 883--921, 2015.

\bibitem{gammell2015batch}
J.~D. Gammell, S.~S. Srinivasa, and T.~D. Barfoot, ``Batch informed trees
  (bit*): Sampling-based optimal planning via the heuristically guided search
  of implicit random geometric graphs,'' in \emph{2015 IEEE international
  conference on robotics and automation (ICRA)}.\hskip 1em plus 0.5em minus
  0.4em\relax IEEE, 2015, pp. 3067--3074.

\bibitem{strub2020adaptively}
M.~P. Strub and J.~D. Gammell, ``Adaptively informed trees (ait*): Fast
  asymptotically optimal path planning through adaptive heuristics,'' in
  \emph{2020 IEEE International Conference on Robotics and Automation
  (ICRA)}.\hskip 1em plus 0.5em minus 0.4em\relax IEEE, 2020, pp. 3191--3198.

\bibitem{koenig2004lifelong}
S.~Koenig, M.~Likhachev, and D.~Furcy, ``Lifelong planning a*,''
  \emph{Artificial Intelligence}, vol. 155, no. 1-2, pp. 93--146, 2004.

\bibitem{kuffner2000rrt}
J.~J. Kuffner and S.~M. LaValle, ``Rrt-connect: An efficient approach to
  single-query path planning,'' in \emph{Proceedings 2000 ICRA. Millennium
  Conference. IEEE International Conference on Robotics and Automation.
  Symposia Proceedings (Cat. No. 00CH37065)}, vol.~2.\hskip 1em plus 0.5em
  minus 0.4em\relax IEEE, 2000, pp. 995--1001.

\bibitem{mandalika2021guided}
A.~Mandalika, R.~Scalise, B.~Hou, S.~Choudhury, and S.~S. Srinivasa, ``Guided
  incremental local densification for accelerated sampling-based motion
  planning,'' \emph{arXiv preprint arXiv:2104.05037}, 2021.

\bibitem{zhang2018learning}
C.~Zhang, J.~Huh, and D.~D. Lee, ``Learning implicit sampling distributions for
  motion planning,'' in \emph{2018 IEEE/RSJ International Conference on
  Intelligent Robots and Systems (IROS)}.\hskip 1em plus 0.5em minus
  0.4em\relax IEEE, 2018, pp. 3654--3661.

\bibitem{ichter2020learned}
B.~Ichter, E.~Schmerling, T.-W.~E. Lee, and A.~Faust, ``Learned critical
  probabilistic roadmaps for robotic motion planning,'' in \emph{2020 IEEE
  International Conference on Robotics and Automation (ICRA)}.\hskip 1em plus
  0.5em minus 0.4em\relax IEEE, 2020, pp. 9535--9541.

\bibitem{ichter2018learning}
B.~Ichter, J.~Harrison, and M.~Pavone, ``Learning sampling distributions for
  robot motion planning,'' in \emph{2018 IEEE International Conference on
  Robotics and Automation (ICRA)}.\hskip 1em plus 0.5em minus 0.4em\relax IEEE,
  2018, pp. 7087--7094.

\bibitem{doersch2016tutorial}
C.~Doersch, ``Tutorial on variational autoencoders,'' \emph{arXiv preprint
  arXiv:1606.05908}, 2016.

\bibitem{kumar2019lego}
R.~Kumar, A.~Mandalika, S.~Choudhury, and S.~Srinivasa, ``Lego: Leveraging
  experience in roadmap generation for sampling-based planning,'' in \emph{2019
  IEEE/RSJ International Conference on Intelligent Robots and Systems
  (IROS)}.\hskip 1em plus 0.5em minus 0.4em\relax IEEE, 2019, pp. 1488--1495.

\bibitem{jenamani2020robotic}
R.~K. Jenamani, R.~Kumar, P.~Mall, and K.~Kedia, ``Robotic motion planning
  using learned critical sources and local sampling,'' \emph{arXiv preprint
  arXiv:2006.04194}, 2020.

\bibitem{qureshi2018deeply}
A.~H. Qureshi and M.~C. Yip, ``Deeply informed neural sampling for robot motion
  planning,'' in \emph{2018 IEEE/RSJ International Conference on Intelligent
  Robots and Systems (IROS)}.\hskip 1em plus 0.5em minus 0.4em\relax IEEE,
  2018, pp. 6582--6588.

\bibitem{qureshi2020motion}
A.~H. Qureshi, Y.~Miao, A.~Simeonov, and M.~C. Yip, ``Motion planning networks:
  Bridging the gap between learning-based and classical motion planners,''
  \emph{IEEE Transactions on Robotics}, vol.~37, no.~1, pp. 48--66, 2020.

\bibitem{khan2020graph}
A.~Khan, A.~Ribeiro, V.~Kumar, and A.~G. Francis, ``Graph neural networks for
  motion planning,'' \emph{arXiv preprint arXiv:2006.06248}, 2020.

\bibitem{kampffmeyer2018connnet}
M.~Kampffmeyer, N.~Dong, X.~Liang, Y.~Zhang, and E.~P. Xing, ``Connnet: A
  long-range relation-aware pixel-connectivity network for salient
  segmentation,'' \emph{IEEE Transactions on Image Processing}, vol.~28, no.~5,
  pp. 2518--2529, 2018.

\bibitem{turaga2009maximin}
S.~C. Turaga, K.~L. Briggman, M.~Helmstaedter, W.~Denk, and H.~S. Seung,
  ``Maximin affinity learning of image segmentation,'' \emph{arXiv preprint
  arXiv:0911.5372}, 2009.

\bibitem{funke2018large}
J.~Funke, F.~Tschopp, W.~Grisaitis, A.~Sheridan, C.~Singh, S.~Saalfeld, and
  S.~C. Turaga, ``Large scale image segmentation with structured loss based
  deep learning for connectome reconstruction,'' \emph{IEEE transactions on
  pattern analysis and machine intelligence}, vol.~41, no.~7, pp. 1669--1680,
  2018.

\bibitem{cousty2018hierarchical}
J.~Cousty, L.~Najman, Y.~Kenmochi, and S.~Guimar{\~a}es, ``Hierarchical
  segmentations with graphs: quasi-flat zones, minimum spanning trees, and
  saliency maps,'' \emph{Journal of Mathematical Imaging and Vision}, vol.~60,
  no.~4, pp. 479--502, 2018.

\bibitem{teb_planner}
C.~Rösmann, F.~Hoffmann, and T.~Bertram, ``Kinodynamic trajectory optimization
  and control for car-like robots,'' in \emph{2017 IEEE/RSJ International
  Conference on Intelligent Robots and Systems (IROS)}, 2017, pp. 5681--5686.

\bibitem{hauser2010fast}
K.~Hauser and V.~Ng-Thow-Hing, ``Fast smoothing of manipulator trajectories
  using optimal bounded-acceleration shortcuts,'' in \emph{2010 IEEE
  international conference on robotics and automation}.\hskip 1em plus 0.5em
  minus 0.4em\relax IEEE, 2010, pp. 2493--2498.

\end{thebibliography}

\begin{IEEEbiography}[{\includegraphics[width=1in,height=1.25in,clip,keepaspectratio]{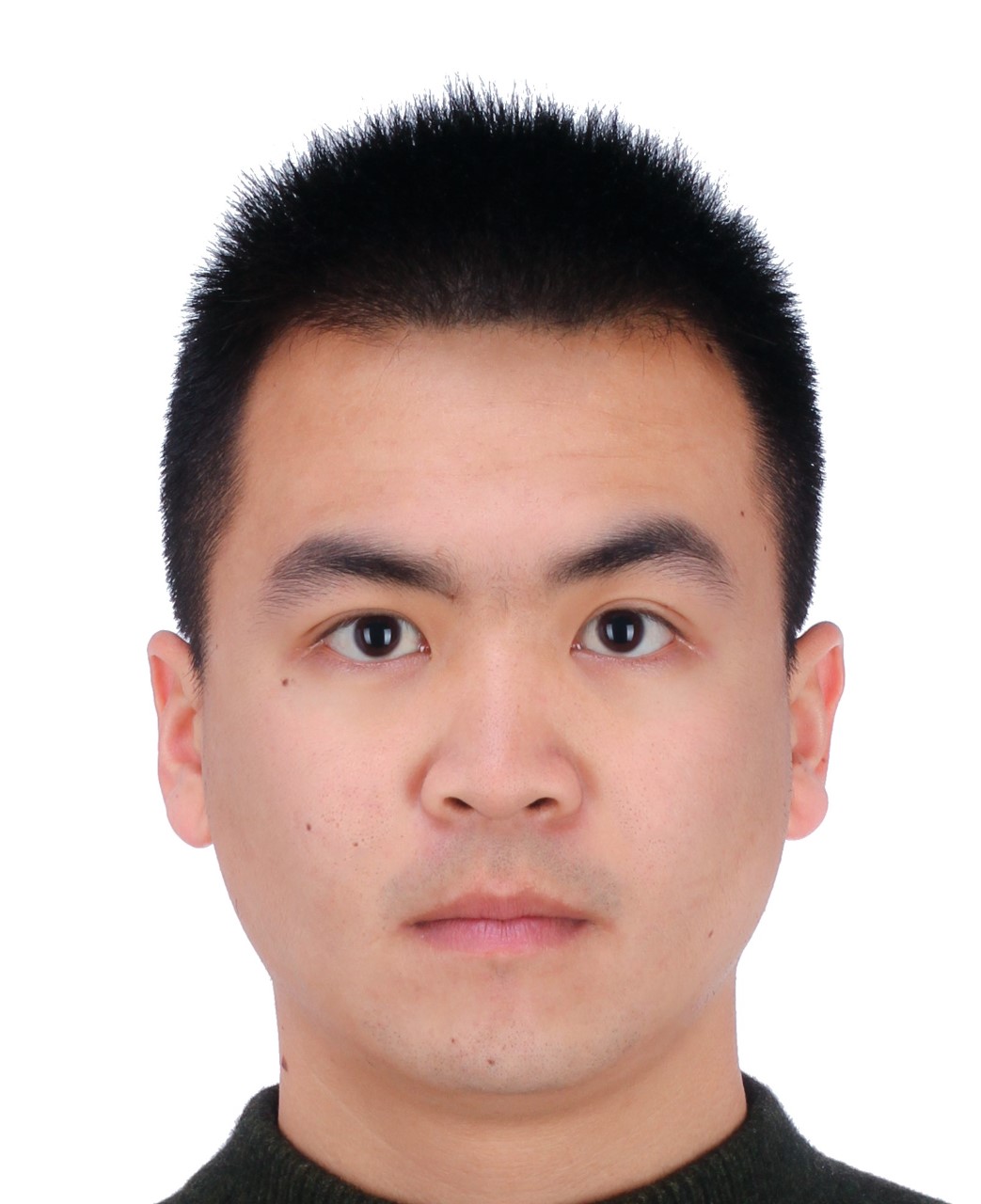}}]{Han Ma}
    received the B.E. degree in measurement, control technology and instrument from the Department of Precision Instrument of Tsinghua University, Beijing, China, in 2019. He is now working towards the Ph.D. degree in the Department of Electronic Engineering of The Chinese University of Hong Kong, Hong Kong SAR, China. His research interests include path planning and machine learning in robotics.
\end{IEEEbiography}

\begin{IEEEbiography}[{\includegraphics[width=1in,height=1.25in,clip,keepaspectratio]{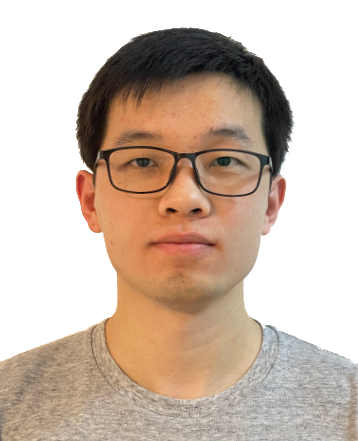}}]{Chenming Li}
    received the B.E. degree in petroleum engineering from the China University of Petroleum, Qingdao, China, in 2017, and the M.Sc. degree in electronic engineering from The Chinese University of Hong Kong, Hong Kong SAR, China, in 2018. He is currently pursuing the Ph.D. degree with the Department of Electronic Engineering, The Chinese University of Hong Kong. His current research interests include robot motion planning, and machine learning in robotics.
\end{IEEEbiography}

\begin{IEEEbiography}[{\includegraphics[width=1in,height=1.25in,clip,keepaspectratio]{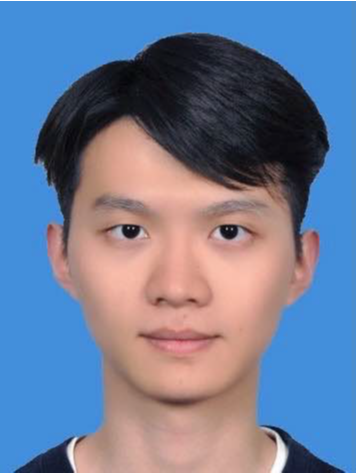}}]{Jianbang Liu}
   received the B.E. degree in microelectronics from Sun Yat-sen University, Guangzhou, China, B.E. degree in electronic engineering from the Hong Kong Polytechnic University , Hong Kong, in 2015, and M.Sc. degree from the University of Hong Kong, Hong Kong in 2016. Currently, he is working toward the Ph.D. degree at the Chinese University of Hong Kong, Hong Kong. His research interests include sensor fusion, path planning, robotic perception for decision making and control.
\end{IEEEbiography}

\begin{IEEEbiography}[{\includegraphics[width=1in,height=1.25in,clip,keepaspectratio]{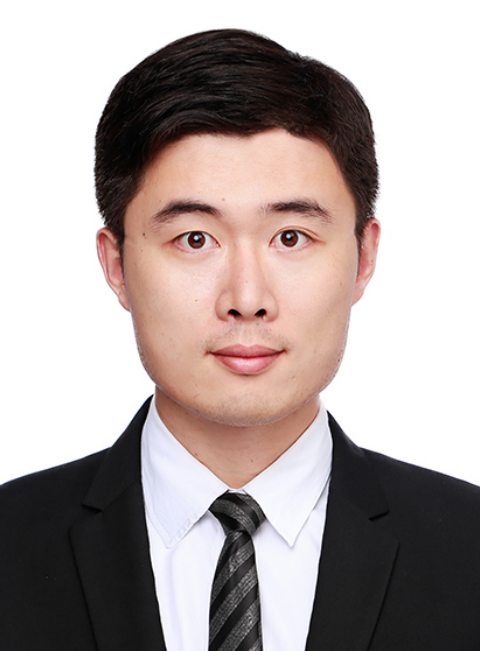}}]{Jiankun Wang}
    received the B.E. degree in Automation from Shandong University, Jinan, China, in 2015, and the Ph.D. degree in Department of Electronic Engineering, The Chinese University of Hong Kong, Hong Kong, in 2019. He is currently an Assistant Professor with the Department of Electronic and Electrical Engineering of the Southern University of Science and Technology, Shenzhen, China.

    During his Ph.D. degree, he spent six months at Stanford University, CA, USA, as a Visiting Student Scholar supervised by Prof. Oussama Khatib. His current research interests include motion planning and control, human robot interaction, and machine learning in robotics.
\end{IEEEbiography}

\begin{IEEEbiography}[{\includegraphics[width=1in,height=1.25in,clip,keepaspectratio]{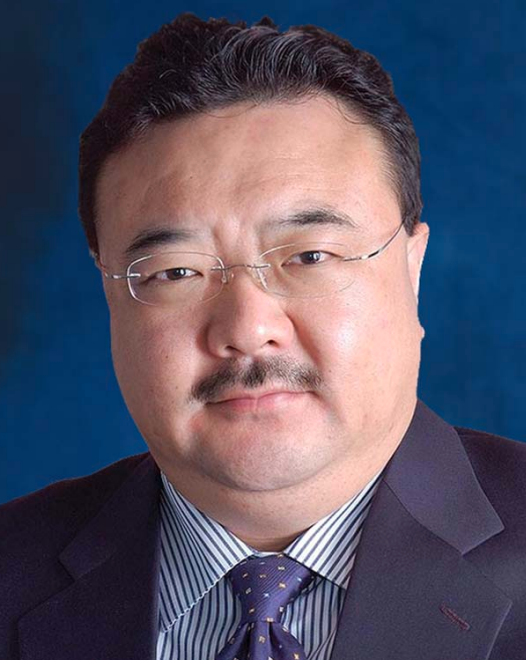}}]{Max Q.-H. Meng}
    received his Ph.D. degree in Electrical and Computer Engineering from the University of Victoria, Canada, in 1992. He is currently a Chair Professor and the Head of the Department of Electronic and Electrical Engineering at the Southern University of Science and Technology in Shenzhen, China, on leave from the Department of Electronic Engineering at the Chinese University of Hong Kong. He joined the Chinese University of Hong Kong in 2001 as a Professor and later the Chairperson of Department of Electronic Engineering. He was with the Department of Electrical and Computer Engineering at the University of Alberta in Canada, where he served as the Director of the ART (Advanced Robotics and Teleoperation) Lab and held the positions of Assistant Professor (1994), Associate Professor (1998), and Professor (2000), respectively. He is an Honorary Chair Professor at Harbin Institute of Technology and Zhejiang University, and also the Honorary Dean of the School of Control Science and Engineering at Shandong University, in China. 
    
    His research interests include medical and service robotics, robotics perception and intelligence. He has published more than 750 journal and conference papers and book chapters and led more than 60 funded research projects to completion as Principal Investigator. 
    
    Prof. Meng has been serving as the Editor-in-Chief and editorial board of a number of international journals, including the Editor-in-Chief of the Elsevier Journal of Biomimetic Intelligence and Robotics, and as the General Chair or Program Chair of many international conferences, including the General Chair of IROS 2005 and ICRA 2021, respectively. He served as an Associate VP for Conferences of the IEEE Robotics and Automation Society (2004-2007), Co-Chair of the Fellow Evaluation Committee and an elected member of the AdCom of IEEE RAS for two terms. He is a recipient of the IEEE Millennium Medal, a Fellow of IEEE, a Fellow of Hong Kong Institution of Engineers, and an Academician of the Canadian Academy of Engineering.

\end{IEEEbiography}

\end{document}